\documentclass[journal]{IEEEtran}

\usepackage{cite}
\usepackage{amsmath,amssymb,amsfonts}
\usepackage{algorithmic}
\usepackage{graphicx}
\usepackage{tabularx}
\usepackage{multirow}
\usepackage{verbatim}
\usepackage[colorlinks=true, allcolors=blue]{hyperref}
\usepackage[table,xcdraw]{xcolor}

\usepackage{orcidlink}

\begin{document}
\title{SDF2Net: Shallow to Deep Feature Fusion Network for PolSAR Image Classification}

\author{Mohammed~Q.~Alkhatib $^{\orcidlink{0000-0003-4812-614X}}$, M.~Sami~Zitouni $^{\orcidlink{0000-0001-7629-8702}}$, Mina~Al-Saad $^{\orcidlink{0000-0003-0393-1700}}$, Nour~Aburaed 
        $^{\orcidlink{0000-0002-5906-0249}}$,  
        and~Hussain~Al-Ahmad $^{\orcidlink{0000-0002-5781-5476}}$
}


\maketitle

\begin{abstract}
Polarimetric synthetic aperture radar (PolSAR) images encompass valuable information that can facilitate extensive land cover interpretation and generate diverse output products. Extracting meaningful features from PolSAR data poses challenges distinct from those encountered in optical imagery. Deep learning (DL) methods offer effective solutions for overcoming these challenges in PolSAR feature extraction. Convolutional neural networks (CNNs) play a crucial role in capturing PolSAR image characteristics by leveraging kernel capabilities to consider local information and the complex-valued nature of PolSAR data. In this study, a novel three-branch fusion of complex-valued CNN, named the Shallow to Deep Feature Fusion Network (SDF2Net), is proposed for PolSAR image classification. To validate the performance of the proposed method, classification results are compared against multiple state-of-the-art approaches using the airborne synthetic aperture radar (AIRSAR) datasets of Flevoland and San Francisco, as well as the ESAR Oberpfaffenhofen dataset. The results indicate that the proposed approach demonstrates improvements in overall accuracy, with a 1.3\% and 0.8\% enhancement for the AIRSAR datasets and a 0.5\% improvement for the ESAR dataset. Analyses conducted on the Flevoland data underscore the effectiveness of the SDF2Net model, revealing a promising overall accuracy of 96.01\% even with only a 1\% sampling ratio. 
Source code is available at:\\ 
\url{ https://github.com/mqalkhatib/SDF2Net}
\end{abstract}

\begin{IEEEkeywords}
complex-valued convolutional neural network (CV-CNN), polarimetric synthetic aperture radar (PolSAR) image classification, Attention Mechanism, Feature Fusion.
\end{IEEEkeywords}

\IEEEpeerreviewmaketitle

\section{Introduction}

\IEEEPARstart{P}{olarimetric}  Synthetic Aperture Radar (PolSAR) images offer a specialized perspective in microwave remote sensing by capturing the polarization properties of radar waves, providing detailed insights into Earth's surface features like vegetation \cite{yin2019optimal}, water bodies \cite{zhang2020automatic}, and man-made structures \cite{xiang2016man}. PolSAR images tackle the limitations of optical remote sensing images, which are susceptible to changes in illumination and weather conditions. Unlike optical systems, PolSAR is capable of functioning in all weather conditions and possesses a robust penetrating capability\cite{shang2022spatial}. Moreover, PolSAR outperforms conventional Synthetic Aperture Radar (SAR) systems by comprehensively capturing extensive scattering information through four different modes. This enables the extraction of comprehensive target details, including scattering echo amplitude, phase, frequency characteristics, and polarization attributes\cite{ren2023new}.

Nowadays, PolSAR imagery has numerous applications in environmental monitoring \cite{brisco2020hybrid}, disaster management \cite{yamaguchi2012disaster}, military monitoring \cite{hou2019semisupervised, lupidi2020polarimetric}, crop prediction \cite{mandal2020sasya,silva2021multitemporal}, and land cover classification \cite{datcu2023explainable}.

PolSAR image classification is the task to classify pixels into specific terrain categories. It involves analyzing the polarization properties of radar waves reflected from Earth's surface. This classification helps automate the identification and mapping of land cover for many applications. 

Conventional approaches to PolSAR image classification primarily rely on extracting distinctive features through the application of target decomposition theory \cite{yang2022polsar}. The Krogager decomposition model, for instance, segregates the scattering matrix into three components, corresponding to helix, diplane, and sphere scattering mechanisms \cite{krogager1990new}. Another widely used method, the Freeman decomposition \cite{freeman1998three}, dissects the polarimetric covariance matrix into double-bounce, surface, and canopy scattering components. Building upon the Freeman decomposition, \cite{yamaguchi2005four} introduced a fourth scattering component, helix scattering power, which proves more advantageous in the classification of PolSAR images. Additionally, the Cloude decomposition \cite{cloude1996review} stands as a common algorithm for PolSAR image analysis. Despite the popularity of traditional classifiers like Support Vector Machine (SVM) \cite{qin2022target} and decision trees \cite{qi2012novel} for PolSAR classification, challenges arise when dealing with PolSAR targets characterized by complex imaging mechanisms. This complexity often leads to inadequacies in representing these targets using conventional features, resulting in diminished classification accuracy.

Recently, Deep Learning (DL) technology has shown remarkable effectiveness in PolSAR image classification \cite{wang2019review, parikh2020classification}. Specifically, Convolutional Neural Networks (CNNs) have exhibited impressive performance in this domain \cite{zhou2016polarimetric, shang2022spatial}. 
Chen et al. \cite{chen2018polsar} successfully harnessed the roll-invariant features of PolSAR targets and the concealed attributes within the rotation domain to train a deep CNN model. This approach contributes to an enhanced classification performance. 
Zhou et al. \cite{zhou2016polarimetric} derived high-level features from the coherency matrix using a deep neural network comprising two convolutional and two fully connected layers, specifically tailored for the analysis of PolSAR images. 
Radman et al. \cite{radman2022dual} discussed the fusion of mini Graph Convolutional Network (miniGCN) and CNN for PolSAR image analysis. Spatial features from Pauli RGB and Yamaguchi were fed into CNN, and polarimetric features were utilized in miniGCN. The study aims to address the limitations of traditional PolSAR image classification methods and presents a dual-branch architecture using miniGCN and CNN models.
Dong et al. \cite{dong2020polsar}  introduces two lightweight 3D-CNN architectures for fast PolSAR interpretation during testing. It applies two lightweight 3D convolution operations and global average pooling to reduce redundancy and computational complexity. The focus is on improving the fully connected layer, which occupies over 90\% of model parameters in CNNs. The proposed architectures use spatial global average pooling to address the computational difficulties and over-fitting risks associated with a large number of parameters.

Phase information, a unique trait to SAR imagery, plays a vital role in various applications like object classification and recognition. Several research studies have explored the application of Complex-valued (CV) CNNs (or simply CV-CNNs) for PolSAR data classification Owing to the difficulties encountered in this domain \cite{barrachina2022real, asiyabi2022complex, hansch2010complex}. Unlike traditional CNNs, CV-CNNs handle complex-valued input data using complex-valued filters and activations, enabling effective capture of both phase and amplitude information. This capability is crucial for accurate PolSAR data classification.

Training CNNs usually demands a substantial volume of data, and acquiring an ample amount of high-quality ground reference data can be both costly and intricate. The scarcity of training samples often results in the generation of unreliable parameters and the risk of overfitting. Hence, it becomes imperative to attain commendable classification performance with a limited set of training data.

Shallow networks excel at capturing simple features but struggle when confronted with more intricate ones. On the contrary, deep neural network structures exhibit proficiency in extracting complex features. The integration of information from various depths in a network facilitates efficient learning, even when working with a modest number of training samples. This integration significantly enhances the network's capability to comprehend the intricate characteristics of the dataset.

In recent years, attention-based techniques have also been widely employed in POlSAR image classification to enhance the model's ability to emphasize informative features and suppress less relevant ones by allocating more attention to the most important features rather than treating the entire input uniformly which in turn improve the classification performance, \cite{fang2023hybrid}.

The main contributions of this paper are demonstrated as follows:
\begin{enumerate}
    \item A novel model is suggested, integrating feature extraction at various depths to enhance the classification performance of PolSAR images effectively.

    \item A feature-learning network with multiple depths and varying layers in each stream is developed. This design enables filters to simultaneously capture shallow, medium, and deep properties, enhancing the utilization of complex information in PolSAR. Experimental results indicate that the proposed model exhibits superior feature-learning capabilities compared to existing models in use.

    \item The model we propose surpasses current methods not only when dealing with a limited number of samples but also attains higher accuracy with an ample training dataset. This conclusion is drawn from statistical outcomes obtained through thorough trials on three PolSAR datasets, which will be elaborated and discussed in the subsequent sections.

\end{enumerate}
The article's structure is outlined as follows: In Section \ref{sec:related}, a concise introduction to related work is provided. Section \ref{sec:Framework} provides a detailed presentation of the proposed network SDF2Net. The experimental results and analysis on three PolSAR datasets are presented in Section \ref{sec:results}. The article concludes with a summary of conclusions and an outlook on future work in Section \ref{sec:conclusion}.

\section{Related Work}
\label{sec:related}
This section offers a concise examination of the relevant literature on CNNs and Attention Mechanisms, with a particular focus on the squeeze and excitation module.

\subsection{Overview CNNs} \label{subsec: cvcnn}
\begin{figure*}[tb!]
\centerline{\includegraphics[width=.90\linewidth]{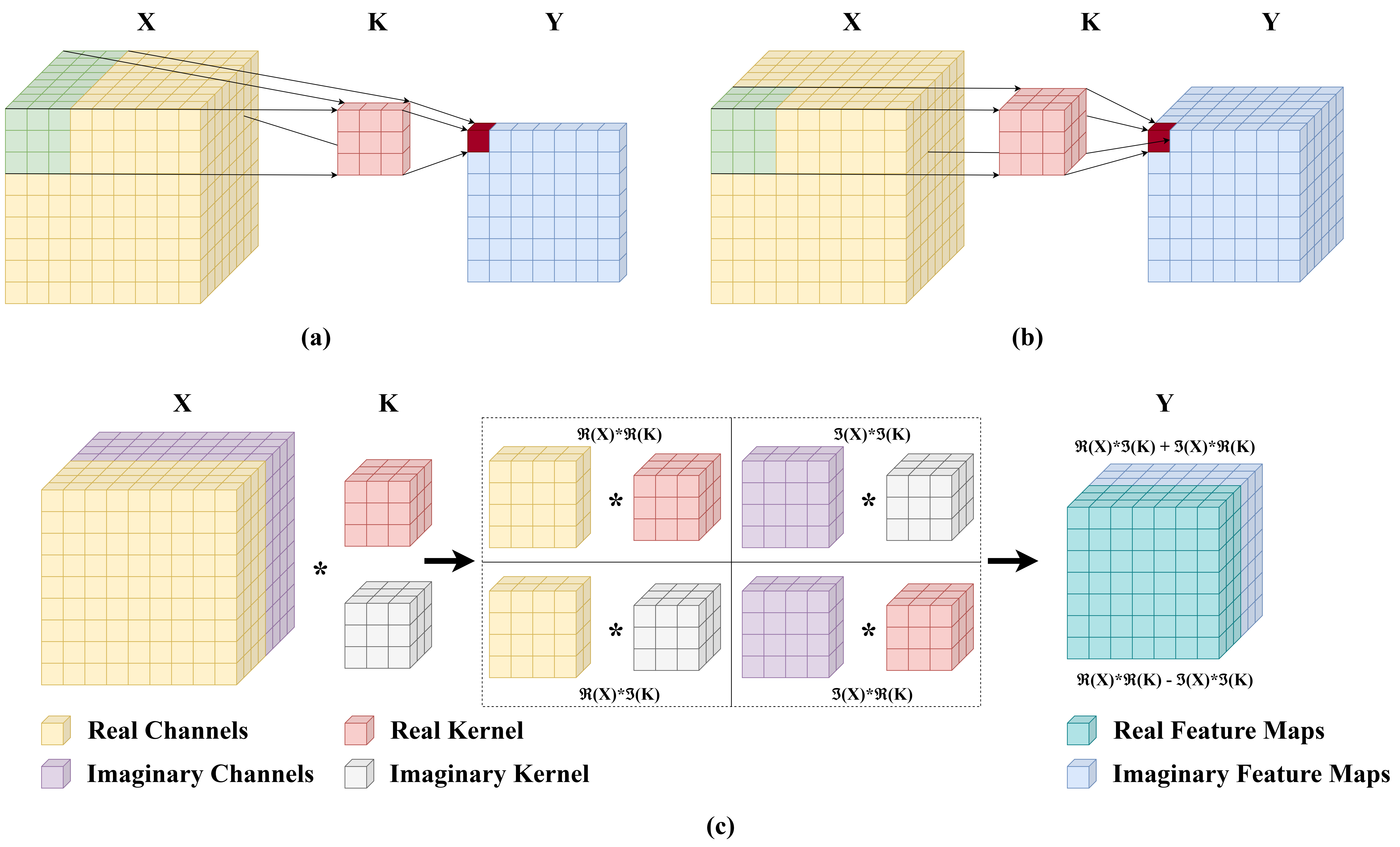}}
\caption{Illustration of different types of convolution on images with multiple channels. (a) 2D Convolution; (b) 3D Convolution; (c) Complex Valued 3D Convolution.}
\label{fig:conv_all}
\end{figure*}

A standard (CNN) consists of an input layer, convolutional layers, activation layer, and an output layer. The initial input layer receives features from the image. Subsequently, convolutional layers employ convolutional kernels (illustrated in Fig. \ref{fig:conv_all}) to extract input features. These kernels operate by taking into account neighboring pixels, considering spatially correlated pixels within a close range (as depicted by the $3\times3$ grid in Fig. \ref{fig:conv_all}(a)). This approach enhances the network's capacity to capture spatially related features. However, 2D-CNNs process each band individually and fail to extract the polarimetric information provided by PolSAR images. To better process PolSAR images, researchers often turn to more advanced architectures, such as Three dimensional (3D) CNNs, or 3D-CNNs.

The application of convolution operations can be expanded into three dimensions, wherein calculations are conducted across all channels concurrently instead of handling each channel separately. Fig. \ref{fig:conv_all}(b) visually depicts the concept of 3D convolution. In 3D convolution, the process encompasses height, width, and channels, making it a suitable approach for incorporating channel context. For an image represented as $X$ with dimensions $N \times N \times B$ and a kernel denoted as $K$ with dimensions $M \times M \times C$, the expression for 3D convolution at position $(x, y, z)$ is given by the following equation:

\begin{equation}
\label{eq:3dcnn}
F_{(x,y,z)} = \sum^M_{i=1}\sum^M_{j=1}\sum^C_{k=1} K_{(i,j,k)}X_{(x+i,y+j, z+k)} + b
\end{equation}


Though 3D-CNNs capture features in a better fashion when compared to 2D-CNNs \cite{zhang2018polarimetric}, the complex nature of PolSAR images, represented by complex-valued data, adds an extra layer of complexity when considering the use of 3D-CNNs for their processing. Compared to traditional CNNs, CV-CNNs proven their superiority in PolSAR classification tasks. To fully explore the complex values in PolSAR data, we utilized CV-3D-CNN. For an image $X = \Re{(X)} + i\Im{(X)}$ and a kernel $k = \Re{(K)} + i\Im{(K)}$, the result of the complex convolution $Y$ can be expressed as:
\begin{multline}
     Y = X \ast K = \Re{(X)} \ast \Re{(K)} - \Im{(X)} \ast \Im{(K)} \\
     + i.(\Re{(X)} \ast \Im{(K)}) + i.(\Im{(X)} \ast \Re{(K)}) 
\end{multline}

 \noindent where $\Re{(.)}$ and $\Im{(.)}$ present the real and the imaginary parts of a CV number, respectively. $i$ is the imaginary number, which value is $\sqrt{-1}$. $Y$ could be expressed as $\Re{(Y)} + i.\Im{(Y)}$. This is also illustrated in Fig. \ref{fig:conv_all}(c).

\subsection{Attention Mechanism}
\begin{figure}[t!]
\centering
\includegraphics[width=.95\linewidth]{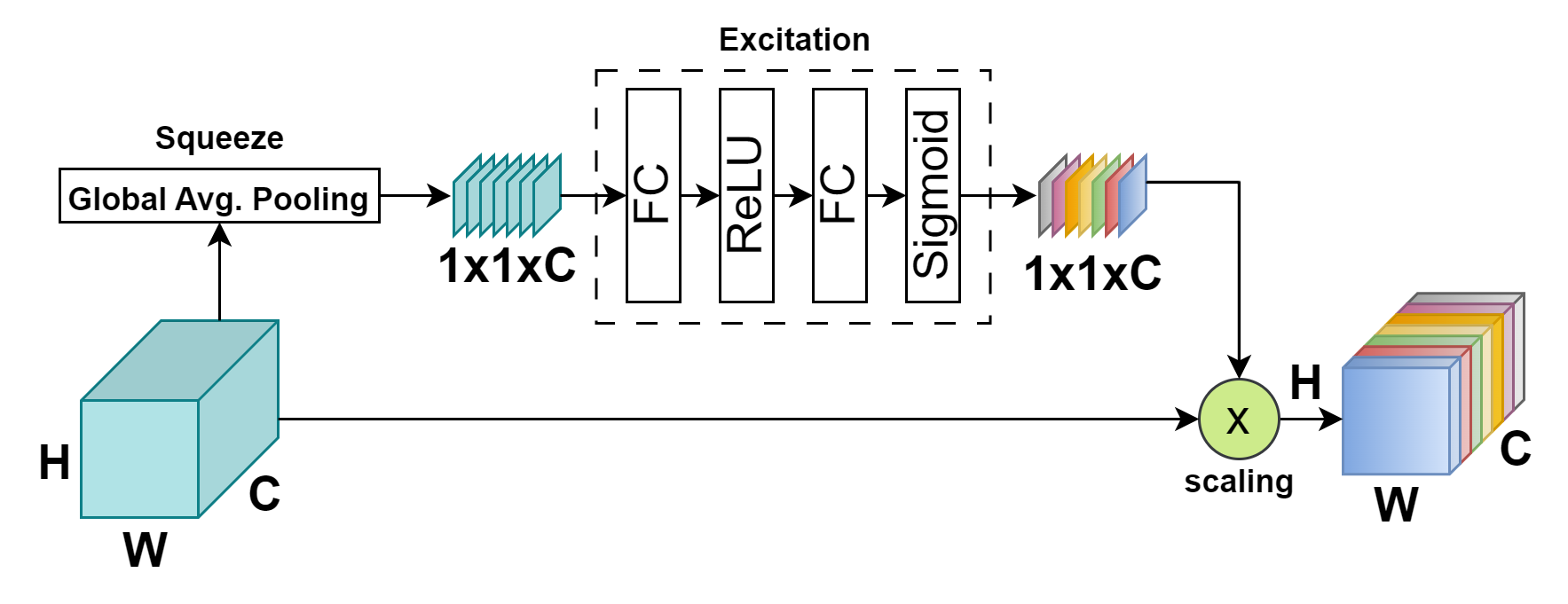}    
 \caption{Squeeze and Excitation Block.}
 \label{fig:SE}
\end{figure}

While CNNs have demonstrated promising outcomes in the classification of PolSAR images, there hasn't been much work regarding the enhancement of input identifiability when developing CNNs. To address this issue, Authors on \cite{dong2020attention} proposed the use of Squeeze and Excitation (SE) \cite{hu2018squeeze} to enhance the performance of CNN by selectively emphasizing crucial features in the input data. Also, SE  improves channel interdependencies with almost no additional computational cost. Fig. \ref{fig:SE} shows the block diagram of SE.

In the SE process \cite{zhang2022sem}, given any transformation (e.g., convolution) $F_{tr}$ that maps the input $X$ to feature maps $u_{c}$, where $u_{c} \in \mathbb{R}^{H\times W\times C}$, and $u_{c}$ represents the c-th $H \times W$ matrix in $u$, with the subscript $c$ denoting the number of channels. In the Squeeze procedure ${F_{sq}(.)}$, global average pooling is employed to transform the input of $H \times W \times C$ into an output of $1 \times 1 \times C$. The squeeze function is defined as

\begin{equation}
\label{Squeeze}
\ z_{c}=F_{sq(u_{c})} = \frac{1}{H\times W}\sum_{i=1}^{H}\!\sum_{j=1}^{W} u_{c}(i,j)\
\end{equation}

The equation reveals that the input data is transformed into a column vector during the squeeze process. The length of this vector corresponds to the number of channels. Subsequently, the excitation operation is conducted to autonomously discern the significance of each feature. This process amplifies features that exert a substantial influence on classification results while suppressing irrelevant features. The excitation function can be expressed as 
\begin{equation}
\label{excitation}
\ s= F_{ex(z,W)} = \sigma(g(z,W))=\sigma(W_{2} 
 ReLU(W_{1}z))\
\end{equation}
Where $\sigma$ represents the Sigmoid activation function, $W_{1}\in \mathbb {R}^{\frac{C}{r}\times C}$ and $W_{2}\in \mathbb {R}^{ C\times \frac{C}{r}}$ denote the two fully connected layers. Here, $W_{1}$ functions as the dimensionality reduction layer with a reduction ratio of $r$, while $W_{2}$ serves as the proportionally identical data-dimensionality increase layer. The variable $z$ corresponds to the output from the preceding squeeze layer, and $W_{1} z$ signifies a fully connected layer operation. The Sigmoid function processes the final output of the excitation process, resulting in a value between zero and one. Following the squeeze and excitation steps, an attention vector $s$ is obtained, which can be employed to adjust the channels of $u$.

\begin{figure*}[t!]
\centering
\includegraphics[width=.98\linewidth]{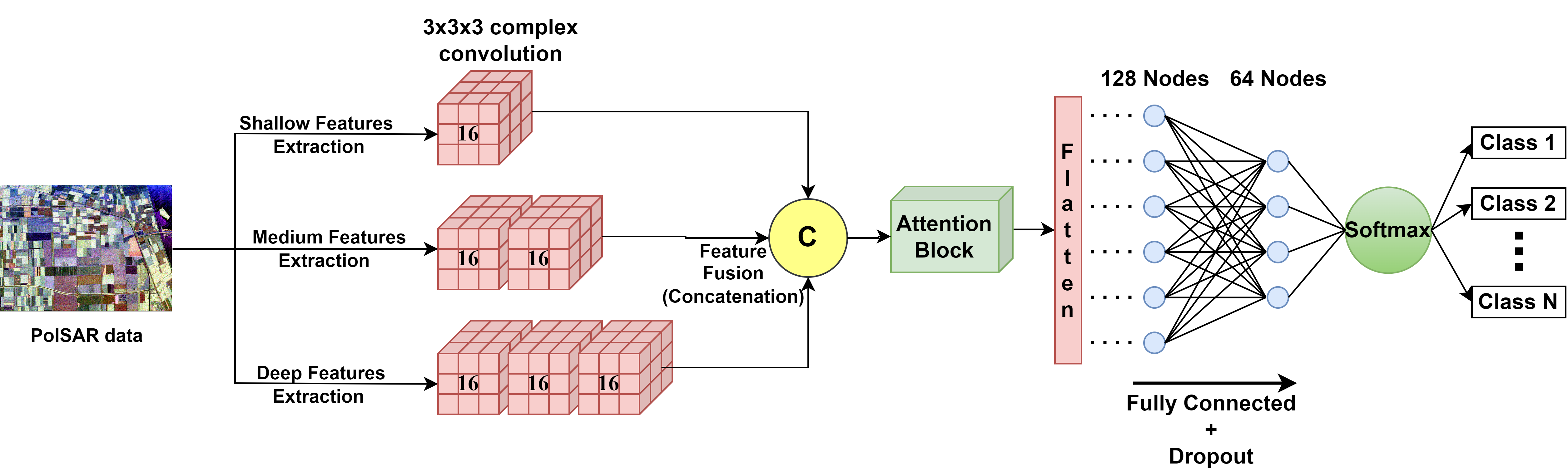}    
 \caption{Block diagram of the proposed SDF2Net.}
 \label{fig:SDF2Net}
\end{figure*}

\section{Methodology} 
\label{sec:Framework}
Within this section, a comprehensive description of the SDF2Net architecture is provided. Initially, the processing of polarimetric data from PolSAR images is showcased, followed by an exposition of the SDF2Net network architecture.

\subsection{PolSAR Data Preprocessing}
The construction of a polarimetric feature vector serves as a fundamental step in PolSAR imagery classification process. In PolSAR imagery, the description of each pixel is defined by a $2 \times 2$ complex scattering matrix, denoted as $S$ as given in Equation \ref{eq:scattering
matrix} \cite{ni2022dnn}

\begin{equation}\label{eq:scattering
matrix}
S = 
\begin{bmatrix}
S_{HH} & S_{HV}  \\
S_{VH} & S_{VV}  
\end{bmatrix}
,
\end{equation}

where $S_{AB}(A,B \in {H,V})$ represents the backscattering coefficient of the polarized electromagnetic wave in emitting $A$ direction and receiving $B$ direction. $H$ and $V$ represent the horizontal and vertical polarization channels, respectively. In the context of data acquired by a monostatic PolSAR radar system, the assumption $S_{VH}$=$S_{HV}$ holds, indicating that the scattering matrix $S$ is symmetric. This enables the simplification and reduction of the matrix to the polarization scattering vector $\vec{k}$. Employing the Pauli decomposition method, the expression for t$\vec{k}$ can be represented as \cite{ren2023new}

\begin{equation}
    \vec{k} = \frac{1}{\sqrt{2}} [S_{HH} + S_{VV} , S_{HH} - S_{VV}, 2S_{HV}]^T.
\end{equation}

In general, multi-look processing is essential for PolSAR data. Following the processing step, the obtained coherency matrix serves as the most commonly used representation for PolSAR data as given in Equation \ref{eq:Hermitian matrix} \cite{ren2023new}
\begin{equation}\label{eq:Hermitian matrix}
    T = \frac{1}{n}\sum_{j = 1}^n \vec{k_j}{\vec{k_j}}^H= \begin{bmatrix}
T_{11} & T_{12}&T_{13}  \\
T_{21} & T_{22}&T_{23}\\
T_{31} & T_{32}&T_{33}
\end{bmatrix},
\end{equation}
where the operator $^H$ stands for complex conjugate operation
and $n$ is the number of looks. It is worth mentioning that $T$ is a Hermitian matrix with real-valued elements on the diagonal and complex-valued elements off-diagonal. As a result, the three real-valued and three complex-valued elements of the upper triangle of the coherency matrix (i.e. $T11$, $T12$, $T13$, $T22$, $T23$, $T33$) are used as the input features of the models.

Data preprocessing is an essential step for achieving higher classification accuracy. Initially, the mean and standard deviation of each channel are computed. Subsequently, normalization is applied to each channel, ensuring optimal data preparation for classification. By taking $T_{11}$ as an example, each channel will be normalized as in Equation \ref{eq:normalize} :

\begin{equation}
    T_{11} = \frac{T_{11} - {\overline{T}}_{11}}{T_{{11}_{std}}}
        ,
    \label{eq:normalize}
\end{equation}
where $\overline{T}$ is the mean operation and $T_{std}$ is the standard deviation.

\subsection{Feature Extraction Using CV-3D-CNN}
As shown in section \ref{subsec: cvcnn}, CV-3D-CNNs are more suitable to process PolSAR data due to their complex nature. Each layer of CV-3D-CNN is followed by a ReLU activation function, and since the output features are complex, we propose the use of Complex Valued ReLu ($\mathbb{C}$ReLU(.)). It is obtained by applying the well-known ReLU(.) to both the real and imaginary parts separately. So that $\mathbb{C}$ReLU(x) = ReLU($\Re$(x)) + \textit{i}.ReLU($\Im$(x)).

\subsection{Architecture of the Proposed SDF2Net}
Currently, the predominant approach in PolSAR classification tasks involves the utilization of 2D-CNN architectures. While 2D-CNNs effectively capture spatial information, they fall short in exploiting the intricate interchannel dependencies inherent in PolSAR images. On the other hand, a three-dimensional convolutional neural network (3D-CNN) exhibits superior feature extraction capabilities when compared to 2D-CNNs \cite{zhang2018polarimetric}.

Fig. \ref{fig:SDF2Net} shows the framework of the complete process of the proposed method. The model processes the data through a three-branch network, extracting features at different levels (shallow, medium and deep), which are later concatenated. The concatenated features then pass through the Attention block to enhance channel dependencies. The flattening layer is employed to transform the concatenated features into one-dimensional vector.
For final classification, two fully connected layers are employed, incorporating dropout to mitigate overfitting, and a softmax layer for generating the final prediction. 

The detailed distribution of parameters of each layer are shown in Fig. \ref{fig:SDF2Net}. 
The first branch incorporates a single layer of complex 3D-CV-CNN with 16 filters, employing a 3$\times$3$\times$3 kernel size to capture shallow features. In contrast, the second branch focuses on medium features with two CV-3D-CNN layers, each maintaining a 3$\times$3$\times$3 kernel size. Meanwhile, the third branch is tailored to extract deep features through the utilization of three layers of CV-3D-CNN, employing 16 filters at each layer, and configuring the kernel size of each filter as 3$\times$3$\times$3. Notably, due to the information loss caused by the pooling layer, it has not been implemented in this network architecture.

\subsection{Loss Function}
The training of the proposed SDF2Net involves the computation of cross-entropy (CE) loss on the training samples. A softmax classifier is utilized to produce the predicted probability matrix for the samples, as shown below
\begin{equation}
    \hat{y}_l^m = Softmax(|\mathbf{x}_{out}|),
\end{equation}

where $\mathbf{x}_{out}$ is the output of the last fully connected layer, and $|.|$ is the magnitude operator. It is worth noting that $\mathbf{x}_{out}$ is complex, while the value of $|\mathbf{x}_{out}| = \sqrt{(\Re{(\mathbf{x}_{out})})^2 + (\Im{(\mathbf{x}_{out})})^2}$ is real, and hence the  value of $  \hat{y}_l^m$ will also be real.
Subsequently, the presentation of the cross-entropy loss $Loss_{CE}$ is depicted as follows
\begin{equation}
Loss_{CE} = -\sum_{m = 1}^M \sum_{l = 1}^L y_l^m \log(\hat{y}_l^m)
\end{equation}
where $y_l^m$ and $\hat{y}_l^m$ are the reference and predicted labels, respectively, $L$ and $M$ are the land cover categories and the overall number of small batch samples, respectively.

\section{Experiments and Results}
\label{sec:results}
In this section, three prevalent datasets in PolSAR classification, namely Flevoland, San Francisco, and Oberpfaffenhofen have been used in our experiments to verify the effectiveness of the proposed approach, coupled with the experimental configurations of the suggested framework.  For a comprehensive demonstration, both visualized classification results and quantitative performance metrics are reported.

\subsection{Polarimetric SAR Datasets}

\begin{figure}[tb!]
\centering
\includegraphics[width=.95\linewidth]{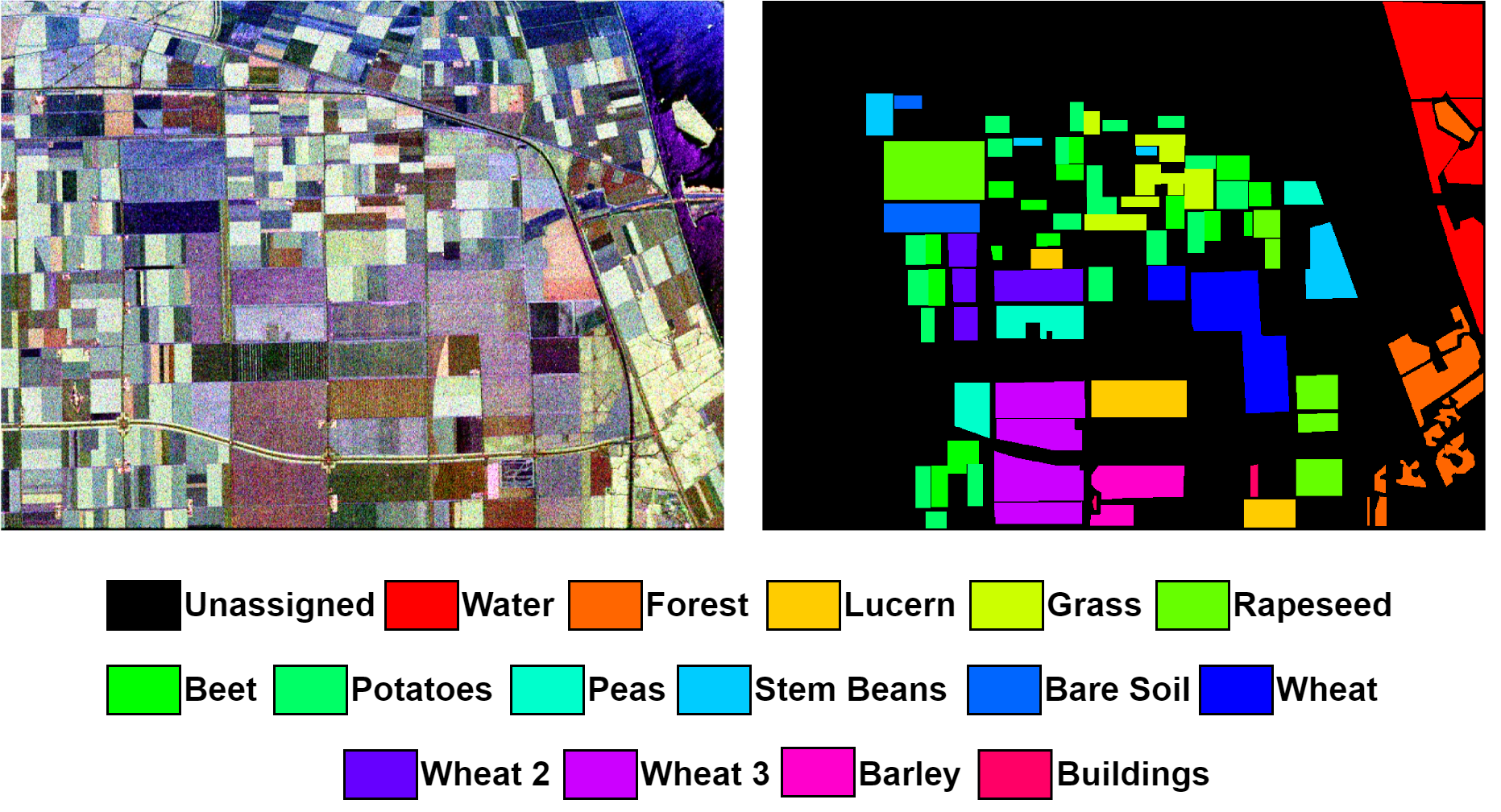}    
 \caption{Flevoland PolSAR data (left) Pauli RGB composite (right) Reference class map.}
 \label{fig:FL_GT}
\end{figure}

\begin{figure}[tb!]
\centering
\includegraphics[width=.95\linewidth]{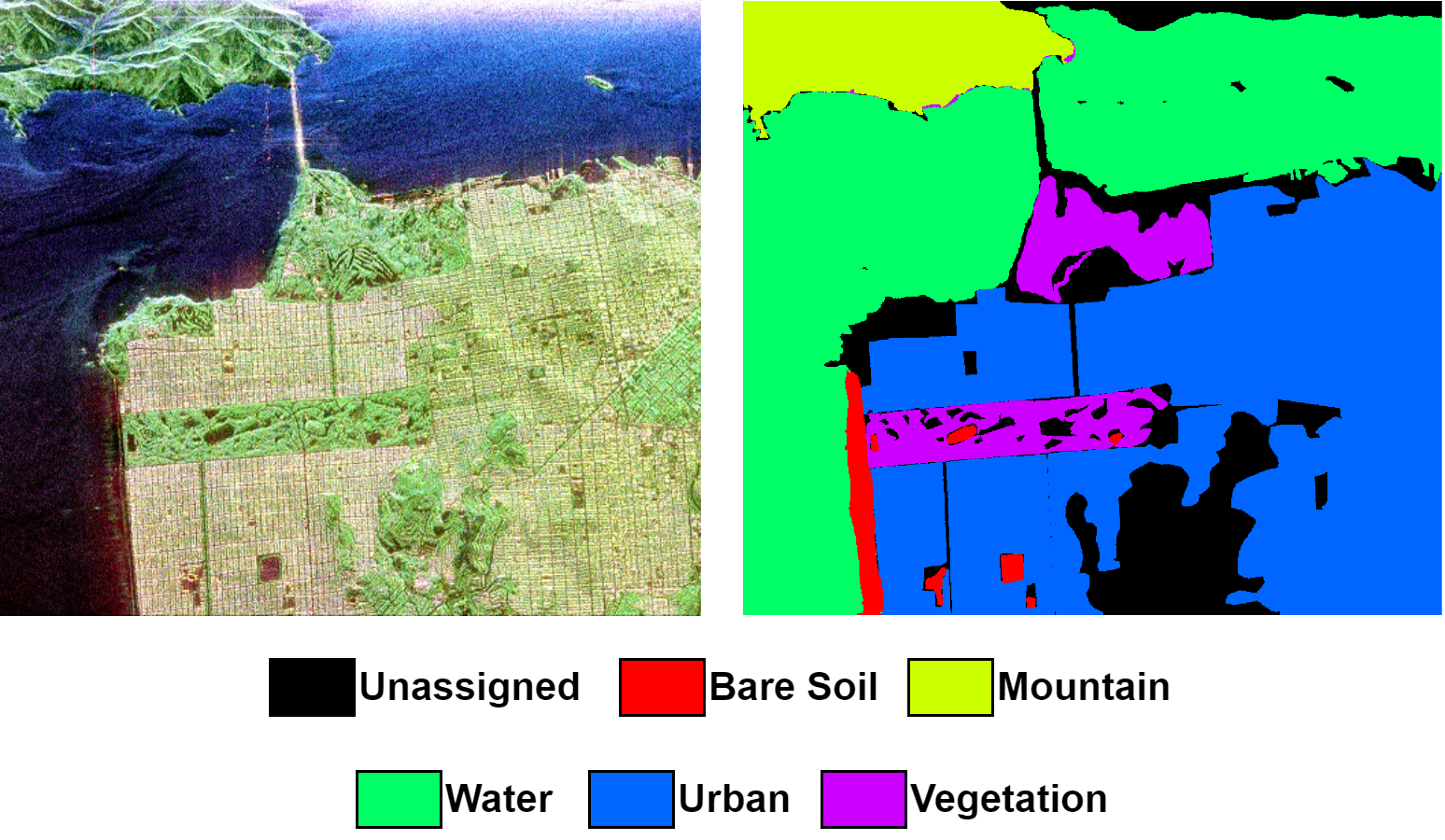}    
 \caption{San Francisco PolSAR data (left) Pauli RGB composite (right) Reference class map.}
 \label{fig:SF_GT}
\end{figure}

\begin{figure}[tb!]
\centering
\includegraphics[width=.95\linewidth]{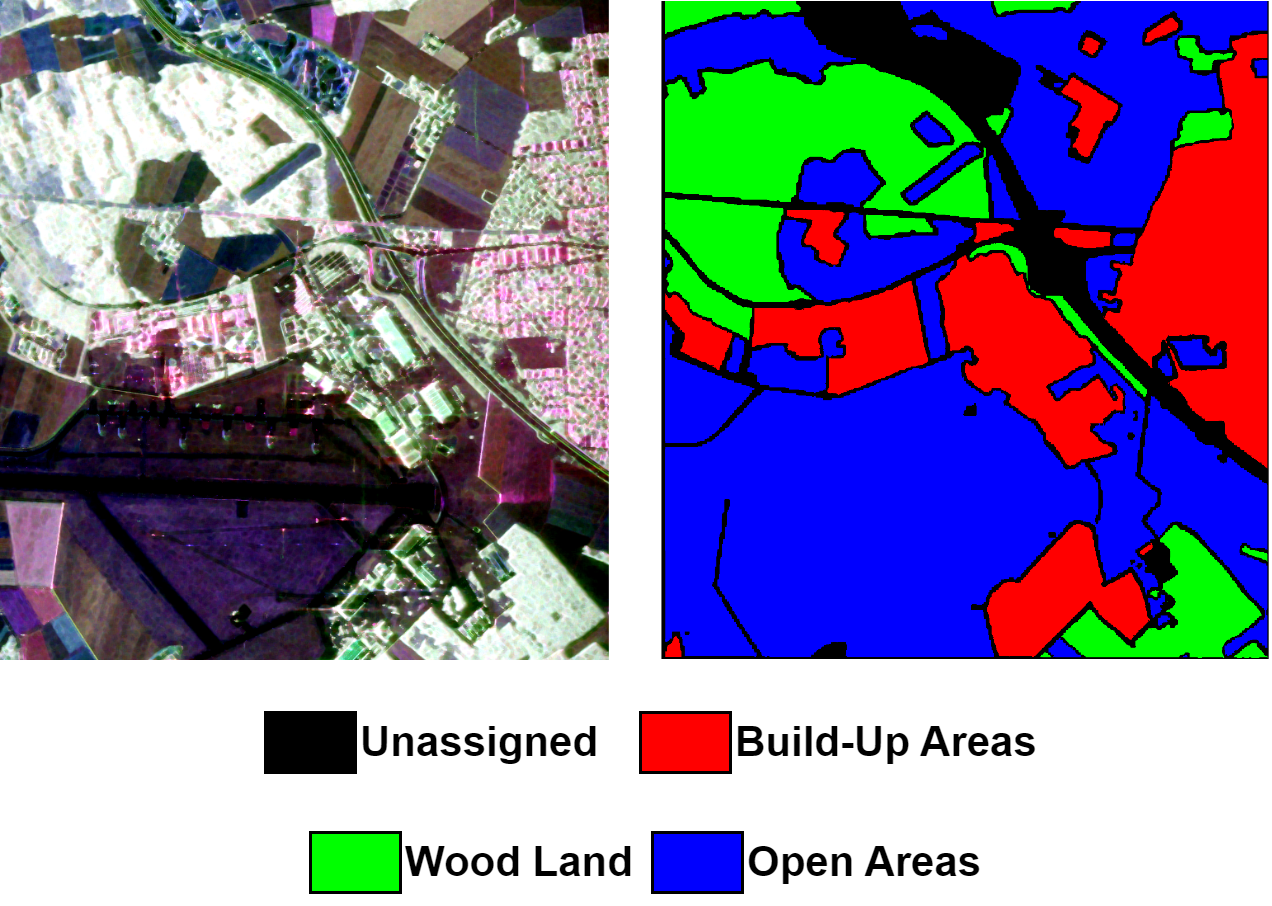}    
 \caption{Oberpfaffenhofen PolSAR data (left) Pauli RGB composite (right) Reference class map.}
 \label{fig:Ober_GT}
\end{figure}

\begin{enumerate}
\item \textbf{Flevoland Dataset}:
The dataset consists of L-band four-look PolSAR data with dimensions 750 $\times$ 1024 pixels with 12 meters. It was acquired by the NASA/JPL AIRSAR system on August 16, 1989 for Flevoland area in the Netherland. It has 15 distinct classes: stem beans, peas, forest, lucerne, wheat, beet, potatoes, bare soil, grass, rapeseed, barley, wheat2, wheat3, water, and buildings\cite{cao2021polsar}. Fig. \ref{fig:FL_GT} shows the Pauli pseudo-color image (Left) and ground truth map (right). Table \ref{tab:flevo} shows the number of pixels per each class in the data set.

\item \textbf{San Francisco Dataset}:
The second dataset, the San Francisco dataset, is obtained from the L-band AIRSAR, covering the San Francisco area in 1989. The image size is 900 $\times$ 1024 pixels and has spatial resolution of 10 meters. It comprises of five categorized terrain classes, encompassing Mountain, Water, Urban, Vegetation and Bare soil \cite{liu2022polsf}. Fig. \ref{fig:SF_GT} shows a color image formed by PauliRGB decomposition to the left, and the reference class map on the right. Table \ref{tab:SF} shows the number of pixels per each class in the data set.

\item \textbf{Oberpfaffenhofen Dataset}:
The Oberpfaffenhofen dataset is captured by L-band ESAR sensor in 2002, encompassing the area of Oberpfaffenhofen in Germany. It includes a PolSAR image with dimensions of 1300 $\times$ 1200 pixels with spatial resolution of 3 meters, annotated with three land cover classes (Built-up Areas, Wood Land, and Open Areas) \cite{hochstuhl2023pol}. Fig. \ref{fig:Ober_GT} shows the PauliRGB composite to the right and the reference class map to the left. Table \ref{tab:ober} shows the number of pixels per each class in the data set.

\end{enumerate}

\begin{table}[tb!]
\centering
\caption{Groundtruth classes for Flevoland scene and their respective samples number.}
\label{tab:flevo}


\begin{tabular}{ccc}
\hline
Class & Name       & Labeled Samples \\ \hline
1     & Water      & 29249           \\
2     & Forest     & 15855           \\
3     & Lucerne    & 11200           \\
4     & Grass      & 10201           \\
5     & Rapeseed   & 21855           \\
6     & Beet       & 14707           \\
7     & Potatoes   & 21344           \\
8     & Peas       & 10396           \\
9     & Stem Beans & 8471            \\
10    & Bare Soil  & 6317            \\
11    & Wheat      & 17639           \\
12    & Wheat 2    & 10629           \\
13    & Wheat 3    & 22022           \\
14    & Barley     & 7369            \\
15    & Buildings  & 578             \\ \hline
Total &            & 207832         
\end{tabular}
\end{table}

\begin{table}[tb!]
\centering
\caption{Groundtruth classes for San Francisco scene and their respective samples number.}
\label{tab:SF}
\begin{tabular}{ccc}
\hline
Class & Name       & Labeled Samples \\ \hline
1     & Bare Soil  & 13701           \\
2     & Mountain   & 62731           \\
3     & Water      & 329566          \\
4     & Urban      & 342795          \\ 
5     & Vegetation & 53509           \\ \hline
Total &            & 802302         
\end{tabular}
\end{table}

\begin{table}[tb!]
\centering
\caption{Groundtruth classes for Oberpfaffenhofen scene and their respective samples number.}
\label{tab:ober}
\begin{tabular}{ccc}
\hline
Class & Name           & Labeled Samples \\ \hline
1     & Build-Up Areas & 328051          \\
2     & Wood Land      & 246673          \\
3     & Open Areas     & 736894          \\ \hline
Total &                & 1311618        
\end{tabular}
\end{table}

\subsection{Evaluation Metrics}

Evaluating the effectiveness of classification results entails comparing predicted class maps with the provided reference or ground truth data. Relying on visual inspection for confirming pixel accuracy in the image is subjective and may lack comprehensiveness. Therefore, a more reliable approach is quantitative evaluation. In this regard, metrics such as Overall Accuracy (OA), Average Accuracy (AA), and the kappa score (k) will be employed.

Overall Accuracy calculates the ratio of correctly assigned pixels to the total number of samples. Average Accuracy computes the mean classification accuracy across all categories or classes. The kappa score assesses the agreement between the predicted classified map and the ground truth, with values ranging from 0 to 1. A value of 1 indicates perfect agreement, while 0 suggests complete disagreement. Typically, a Kappa value equal to or greater than 0.80 signifies substantial agreement, whereas a value below 0.4 indicates poor model performance.

\subsection{Experimental Configuration} 
All the tests were conducted utilizing the Python 3.9 compiler and TensorFlow 2.10.0 framework. Adam optimizer is adopted, where the learning rate is set to $1 \times 10^{-3}$, the batch size is 64 and the training epoch is set to 250. In the course of model training, an early stopping strategy was implemented. Specifically, if there was no improvement in the model's performance over a consecutive span of 10 epochs, the training process was halted, and the model was reverted to its optimal weights. The network configuration of the proposed model using Flevoland dataset is shown in Table \ref{tab:net_config}. The number of samples used for training the model is set to $1\%$ for all three datasets to guarantee a fair comparison.
\begin{table*}[t!]
\centering
\caption{PARAMETER SETTINGS OF SDF2Net on each Dataset with N classes}
\label{tab:net_config}

\resizebox{15cm}{!} {

\begin{tabular}{ccc}

 \hline
\begin{tabular}[c]{@{}c@{}}Shallow Features \\ Extraction Path\end{tabular}                                   & \begin{tabular}[c]{@{}c@{}}Medium Features \\ Extraction Path\end{tabular}                                     & \begin{tabular}[c]{@{}c@{}}Deep Features \\ Extraction Path\end{tabular}                                      \\ \hline
\multicolumn{3}{c}{Input:(13 $\times$ 13 $\times$ 6 $\times$ 1)}                                                                                                                                                                                                                                                                                                    \\ \hline
\begin{tabular}[c]{@{}c@{}}1 $\times$ Complex Convolution\\ (3,3,3,16), Stride = 1, \\ Padding = 'same'\end{tabular} & \begin{tabular}[c]{@{}c@{}}2 $\times$  Complex Convolution\\ (3,3,3,16), Stride = 1, \\ Padding = 'same'\end{tabular} & \begin{tabular}[c]{@{}c@{}}3 $\times$ Complex Convolution\\ (3,3,3,16), Stride = 1, \\ Padding = 'same'\end{tabular} \\ \hline
\multicolumn{1}{l}{Output1:(13 $\times$ 13 $\times$ 6 $\times$ 16)}                                                                & \multicolumn{1}{l}{Output2:(13 $\times$ 13 $\times$ 6 $\times$ 16)}                                                                 & \multicolumn{1}{l}{Output3:(13 $\times$ 13 $\times$ 6 $\times$ 16)}                                                                \\ \hline
\multicolumn{3}{c}{Concat(Output1, Output2, Output3)}                                                                                                                                                                                                                                                                                          \\ \hline
\multicolumn{3}{c}{Output4:(13, $\times$ 13 $\times$ 6 $\times$ 48)}                                                                                                                                                                                                                                                                                                \\ \hline
\multicolumn{3}{c}{Attention Block}                                                                                                                                                                                                                                                                                                            \\ \hline
\multicolumn{3}{c}{Flatten}                                                                                                                                                                                                                                                                                                                    \\ \hline
\multicolumn{3}{c}{Output5:(48,672)}                                                                                                                                                                                                                                                                                                           \\ \hline
\multicolumn{3}{c}{FC-(48,672:128)}                                                                                                                                                                                                                                                                                                            \\ \hline
\multicolumn{3}{c}{Dropout(0.25)}                                                                                                                                                                                                                                                                                                              \\ \hline
\multicolumn{3}{c}{FC-(128:64)}                                                                                                                                                                                                                                                                                                                \\ \hline
\multicolumn{3}{c}{Dropout(0.25)}                                                                                                                                                                                                                                                                                                              \\ \hline
\multicolumn{3}{c}{FC-(64:N)}                                                                                                                                                                                                                                                                                                                 \\ \hline
\multicolumn{3}{c}{Output:(N)}                                                                                                                                                                                                                                                                                                                \\ \hline
\end{tabular}
}
\end{table*}

\subsection{Experimental Results}
In this section, we assess the classification performance of the suggested model both quantitatively and qualitatively, utilizing the three previously mentioned datasets: Flevoland, San Francisco, and Oberpfaffenhofen. To mitigate the impact of sample selection randomness on classification outcomes, the experiments were iterated 10 times, and the final result is presented as the average value of these repetitions. Furthermore, detailed classification outcomes for each category are provided.

\subsubsection{Determining optimal window size}
\begin{figure}[t!]
\centering
\includegraphics[width=.95\linewidth]{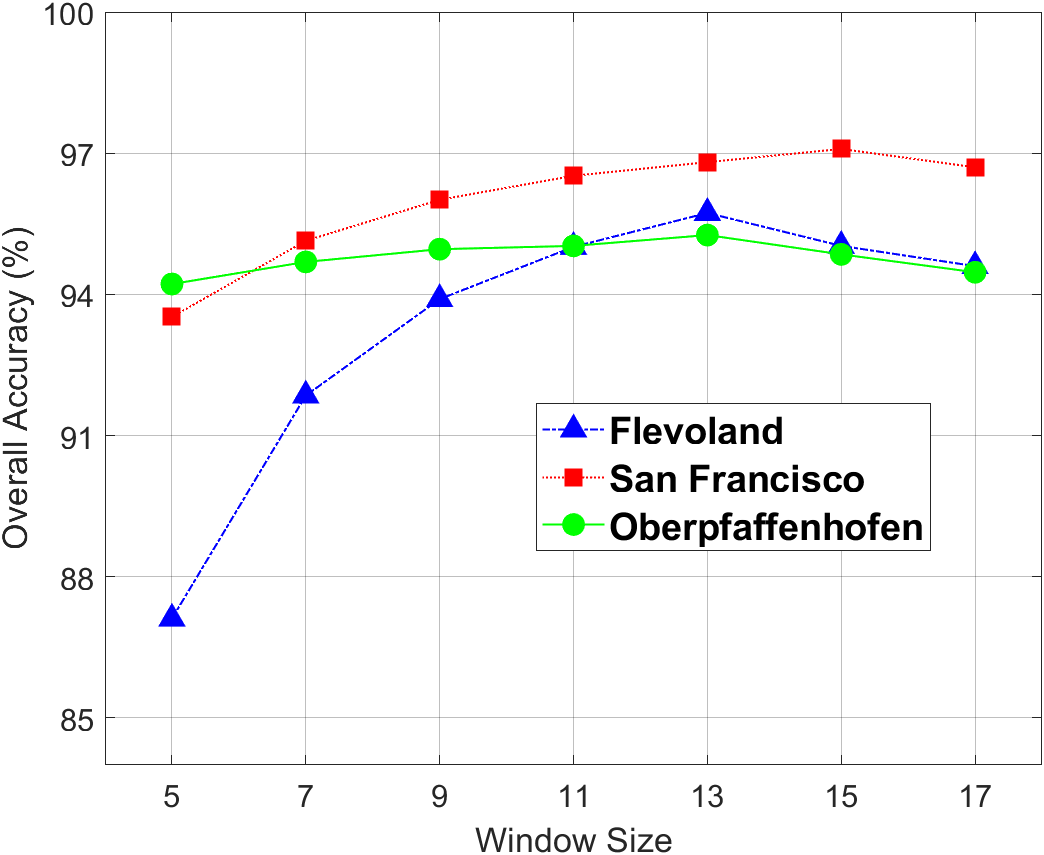}    
 \caption{The overall accuracy of the proposed model employing varying window sizes across the three datasets.}
 \label{fig:optimaWindow}
\end{figure}

In this part, we investigate how the spatial characteristics of diverse datasets influence the proposed model's capability to categorize PolSAR data and identify the optimal window size for each dataset. The window size signifies the extent of spatial information from the retrieved 3D patch that is utilized for assigning a label to the extracted patch. A larger window may encompass a significant amount of neighborhood data, potentially containing information from other classes, thereby impeding the feature extraction process. Conversely, if the chosen window is too small, the model's capacity to extract features will be compromised by a notable loss of spatial information. This study validates the influence of window size on model performance across the three aforementioned datasets. In the experiment, the spatial sizes were configured as  \{$5\times5$, $7\times7$, $9\times9$, $11\times11$, $13\times13$, $15\times15$, $17\times17$\}. It is clear from Fig. \ref{fig:optimaWindow} that for the Flevoland, San Francisco and Oberpfaffenhofen datasets, the suitable window size for the proposed model was $13\times13$, $15\times15$ and $13\times13$, respectively.

\subsubsection{Ablation Study}
Ablation Study: This section is divided into two segments for the ablation study. The initial part focuses on assessing the influence of diverse combinations of network components. We conducted comprehensive experiments on the Flevoland dataset, as outlined in Table \ref{tab:ablation1}. The results demonstrate that our proposed fusion technique outperforms other combinations or fusion methods, such as Shallow (S), Medium (M), Deep (D), Shallow and Medium (S+M), Shallow and Deep (S+D), Medium and Deep (M+D), in terms of AA, OA, and Kappa metrics.
\begin{table}[b!]
\centering
\caption{Results of ablation studies on different combination of model branches over the Flevoland dataset.}
\label{tab:ablation1}
\begin{tabular}{cccc}
\hline
Combination & OA (\%)                   & AA (\%)                   & k $\times$ 100            \\ \hline
S           & 93.01$\pm$0.59          & 91.38$\pm$  0.68        & 92.46$\pm$0.64        \\ 
M           & 93.87$\pm$0.77        & 93.76$\pm$0.54        & 93.49$\pm$0.84        \\ 
D           & 94.69$\pm$0.33        & 93.52$\pm$0.47        & 94.37$\pm$0.36        \\ 
S+M         & 93.35$\pm$0.45         & 93.56$\pm$0.46         & 93.76$\pm$0.63        \\ 
S+D         & 94.71$\pm$0.26        & 93.49$\pm$0.63        & 94.29$\pm$0.36        \\ 
M+D         & 95.57$\pm$0.24        & 94.59$\pm$0.24           & 95.30$\pm$0.37        \\ 
Proposed    & \textbf{96.01$\pm$0.40} & \textbf{95.17$\pm$0.62} & \textbf{95.64$\pm$0.44} \\ \hline
\end{tabular}
\end{table}

 The second part examines the impact of incorporating the attention mechanism. Initially, we experimented with the model without any attention mechanism, followed by placing attention at each stream before feature fusion, and finally applying attention after feature fusion. Table \ref{tab:ablation2} presents the outcomes for various attention locations on the same dataset. It is evident that the optimal location is immediately after the feature fusion step.
\begin{table}[b!]
\centering
\caption{Impact of Attention mechanism over the Flevoland dataset.}
\label{tab:ablation2}
\begin{tabular}{cccc}
\hline
Attention Location & OA (\%)          & AA (\%)          & k $\times$ 100   \\ \hline
Without Attention       & 95.14$\pm$0.26 & 94.27$\pm$0.37 & 94.69$\pm$0.28 \\ 
Before Fusion    & 95.84$\pm$0.21 & 94.86$\pm$0.68 & 94.91$\pm$0.32 \\ 
After Fusion  & 96.01$\pm$0.04 & 95.17$\pm$0.62 & 95.64$\pm$0.44 \\ \hline
\end{tabular}
\end{table}

\subsubsection{Comparison with Other Methods}
Several techniques have been chosen for comparison with the SDF2Net model, such as SVM \cite{lardeux2009support}, 2D-CVNN \cite{zhang2017complex}, 3D-CVNN \cite{tan2019complex}, Wavelet CNN \cite{jamali2022polsar}, and our previously proposed method in \cite{alkhatib2023polsar}, namely CV-CNN-SE. The specific experimental configurations for these methods are outlined below.

\begin{table*}[!t]
\centering
\caption{Experimental Results of different methods on Felvoland Dataset.}
\label{tab:FL_Results}
\begin{tabular}{ccccccccc}
\hline
Class      & Train & Test  & SVM              & 2D-CVNN          & Wavelet CNN      & CV-CNN-SE           & 3D-CVNN          & SDF2Net                   \\ \hline
Water      & 292   & 28957 & 81.91            & 97.05            & 99.09            & 99.32            & 99.33            & \textbf{99.81}            \\
Forest     & 159   & 15696 & 71.71            & 81.44            & 85.39            & 98.80            & 95.11            & \textbf{99.23}            \\
Lucerne    & 112   & 11088 & 82.04            & 93.40            & \textbf{98.29}   & 96.32            & 90.48            & 97.46                     \\
Grass      & 102   & 10099 & 0.24             & 5.62             & 83.90            & 86.19            & \textbf{91.57}   & 85.10                     \\
Rapeseed   & 219   & 21636 & 68.99            & 71.88            & 88.25            & 93.87            & \textbf{97.31}   & 94.05                     \\
Beet       & 147   & 14560 & 68.1             & 67.92            & 74.78            & 77.65            & 91.51            & \textbf{91.59}            \\
Potatoes   & 213   & 21131 & 79.4             & 79.11            & \textbf{95.93}   & 95.39            & 94.69            & 91.30                     \\
Peas       & 104   & 10292 & 68.33            & 92.72            & \textbf{99.19}   & 97.65            & 92.76            & 95.80                     \\
Stem Beans & 85    & 8386  & 73.01            & 68.48            & 91.45            & 95.90            & 93.20            & \textbf{99.00}            \\
Bare Soil  & 63    & 6254  & 0                & 0.00             & \textbf{95.06}   & 94.08            & 84.88            & 94.49                     \\
Wheat      & 176   & 17463 & 73.97            & 69.24            & 96.41            & \textbf{98.78}   & 89.55            & 98.16                     \\
Wheat 2    & 106   & 10523 & 0.05             & 22.04            & 72.03            & 81.86            & 95.50            & \textbf{97.34}            \\
Wheat 3    & 220   & 21802 & 83.86            & 95.94            & 97.53            & 98.62            & 97.72            & \textbf{98.86}            \\
Barley     & 74    & 7295  & 0                & 73.08            & 96.51            & 96.76            & 94.19            & \textbf{98.51}            \\
Buildings  & 6     & 572   & 1.04             & 80.97            & 84.60            & 86.33            & \textbf{100.00}  & 86.85                     \\ \hline
OA (\%)    &       &       & 63.22 $\pm$ 0.86 & 73.09 $\pm$ 2.53 & 91.73 $\pm$ 4.15 & 94.78 $\pm$ 1.42 & 94.51 $\pm$ 0.74 & \textbf{96.01 $\pm$ 0.40} \\
AA (\%)    &       &       & 50.18 $\pm$ 0.59 & 66.59 $\pm$ 1.47 & 90.56 $\pm$ 5.30 & 93.17 $\pm$ 2.12 & 93.85 $\pm$ 0.72 & \textbf{95.17 $\pm$ 0.62} \\
k $\times$ 100      &       &       & 59.18 $\pm$ 1.67 & 70.38 $\pm$ 4.31 & 90.96 $\pm$ 5.43 & 93.92 $\pm$ 1.61 & 94.00 $\pm$ 0.79 & \textbf{95.64 $\pm$ 0.44} \\ \hline
\end{tabular}
\end{table*}
\begin{figure*}[t!]
\centering
\includegraphics[width=.95\linewidth]{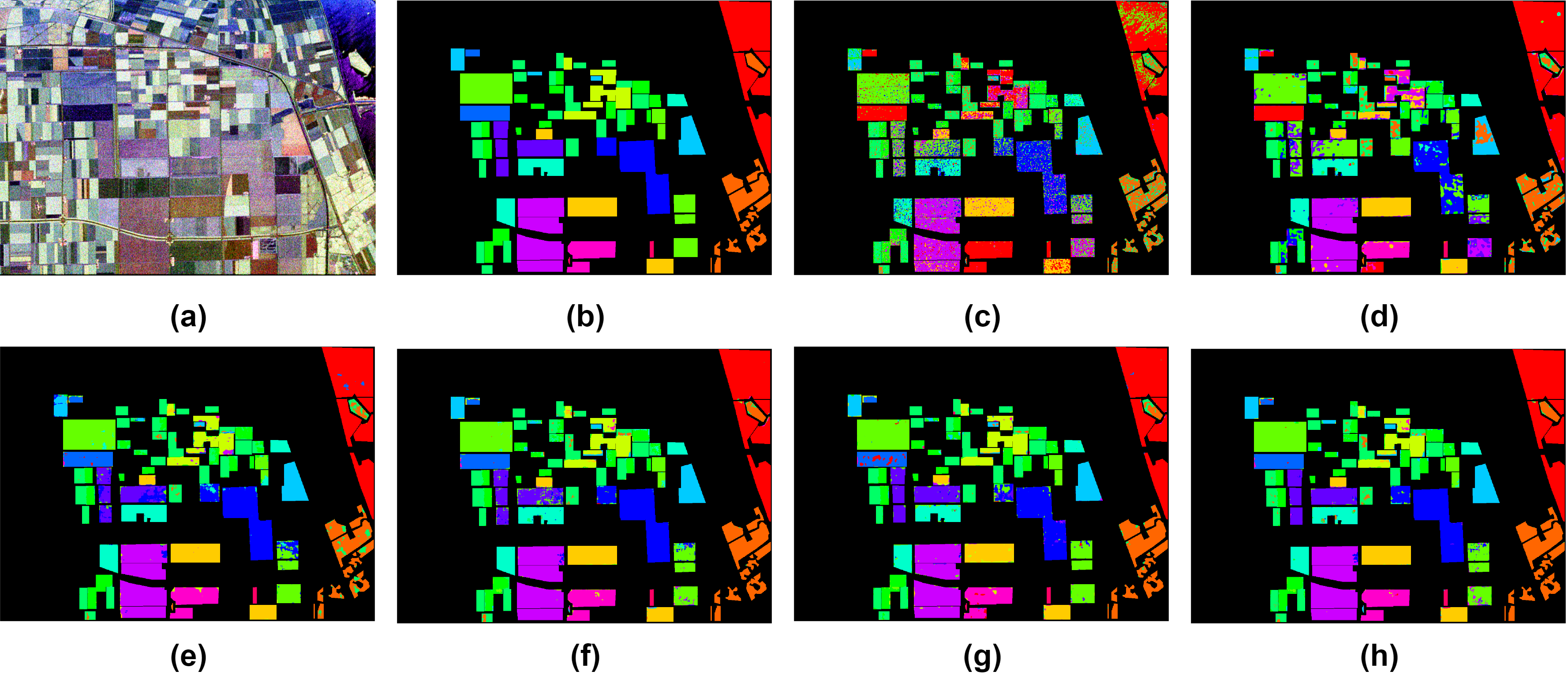}    
 \caption{Classification results of the Flevoland dataset. (a) PauliRGB; (b) Reference Class Map; (c) SVM; (d) 2D-CVNN; (e) Wavelet CNN; (f) CV-CNN-SE; (g) 3D-CVNN; Proposed SDF2Net}
 \label{fig:FL_Results}
\end{figure*}

\begin{itemize}
    \item SVM: The SVM employs the Radial Basis Function (RBF) kernel, with the parameter $\gamma$ set to 0.001 to regulate the local scope of the RBF kernel.

    \item 2D-CVNN: The model consists of two Complex-Valued CNN layers, with 6 and 12 kernels of size $3\times3$ in each layer, and two fully connected layers. The input patch size is specified as $12\times12$ in \cite{zhang2017complex}.

     \item Wavelet CNN: The proposed model utilizes the Haar wavelet transform for feature extraction to improve the classification accuracy of PolSAR imagery. It consists of three branches, each utilizing different concepts and advantages of CNNs. The model parameters were configured based on the values given in \cite{jamali2022polsar}

    \item CV-CNN-SE: This model utilizes the use of 2D-CVNNs at different scales to extract features from PolSAR data. Extracted features are then fused and passed to SE block to enhance classification performance.

    \item 3D-CVNN: In this model, four Complex-Valued CNN layers, with 16, 16, 32 and 32 kernels of size $3\times3\times3$ in each layer, and one fully connected layer. The input patch size is specified as $12\times12$ in \cite{tan2019complex}.

\end{itemize}

As previously stated, the experiments were carried out and iterated 10 times. In each of the 10 trials, only the classification outcomes with the highest accuracy were documented for all algorithms. The quantitative assessments of these compared methods are presented in Tables \ref{tab:FL_Results}--\ref{tab:ober_Results}, with the best results in each table highlighted in bold.

\begin{table*}[t!]
\centering
\caption{Experimental Results of different methods on San Francisco Dataset.}
\label{tab:SF_Results}
\begin{tabular}{ccccccccc}
\hline
Class          & Train & Test   & SVM              & 2D-CVNN          & Wavelet CNN      & CV-CNN-SE           & 3D-CVNN          & SDF2Net                   \\ \hline
Bare Soil      & 137   & 13564  & 0.04             & 47.49            & 78.97            & 57.81            & 73.13            & \textbf{79.98}            \\
Mountain       & 627   & 62104  & 40.61            & 91.27            & 94.62            & 94.82            & \textbf{96.31}   & 94.49                     \\
Water          & 3295  & 326270 & 98.37            & \textbf{99.37}   & 99.26            & 99.11            & 99.24            & 98.70                     \\
Urban          & 3428  & 339367 & 95.65            & 97.78            & 96.21            & 98.47            & 95.52            & \textbf{98.94}            \\
Vegetation     & 535   & 52974  & 64.21            & 78.87            & 60.25            & 77.71            & 86.44   & \textbf{87.42}                     \\ \hline
OA (\%)        &       &        & 88.73 $\pm$ 0.12 & 95.80 $\pm$ 0.37 & 94.65 $\pm$ 2.00 & 96.37 $\pm$ 0.22 & 96.19 $\pm$ 0.32 & \textbf{97.13 $\pm$ 0.20} \\
AA (\%)        &       &        & 59.77 $\pm$ 0.86 & 82.95 $\pm$ 3.23 & 85.86 $\pm$ 4.28 & 85.58 $\pm$ 1.78 & 90.33 $\pm$ 1.3  & \textbf{91.31 $\pm$ 1.54} \\
k $\times$ 100 &       &        & 81.75 $\pm$ 0.21 & 93.38 $\pm$ 0.60 & 91.58 $\pm$ 3.08 & 94.28 $\pm$ 0.35 & 94.07 $\pm$ 0.31 & \textbf{95.50 $\pm$ 0.31} \\ \hline
\end{tabular}
\end{table*}

\begin{figure*}[t!]
\centering
\includegraphics[width=.95\linewidth]{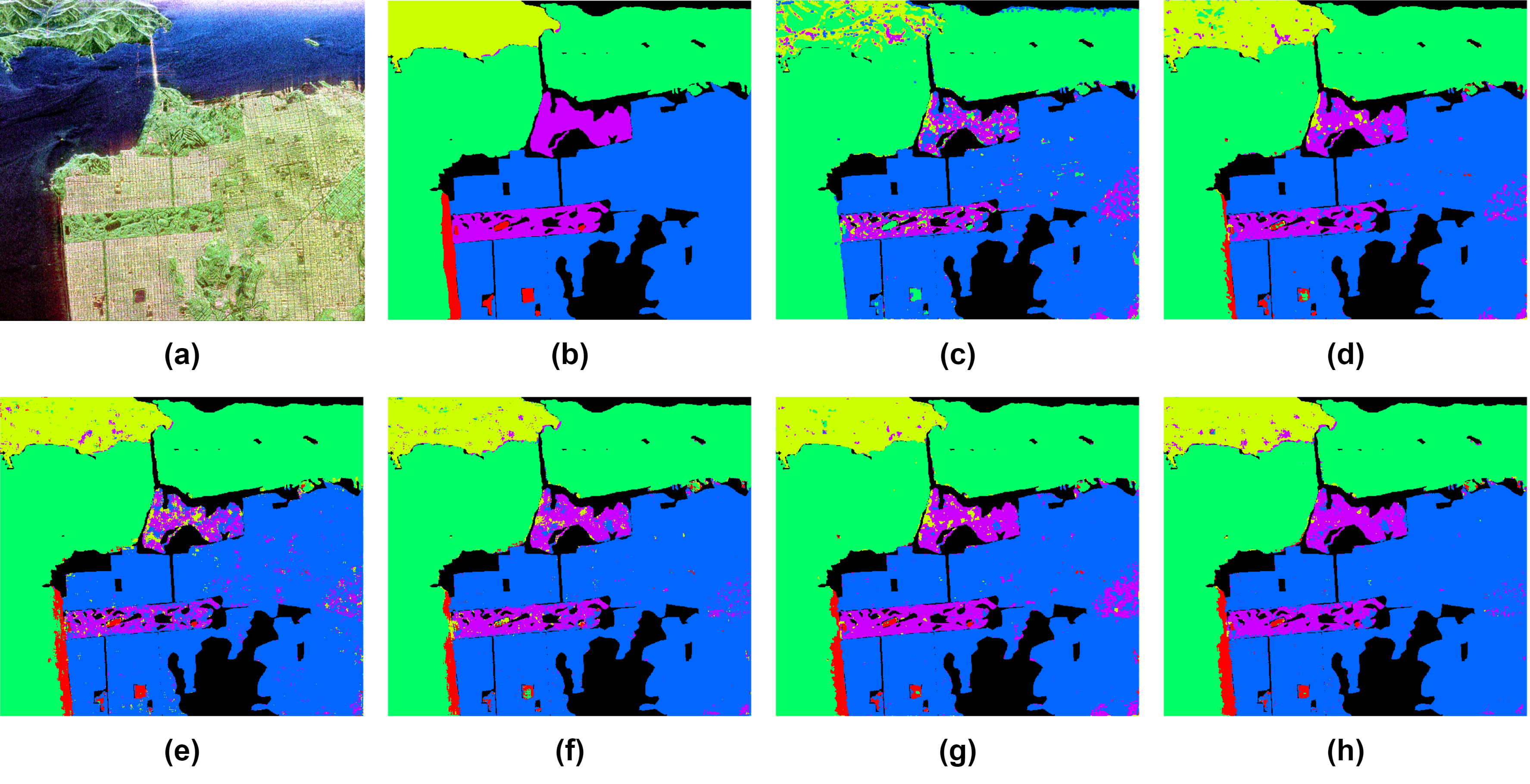}    
 \caption{Classification results of the San Francisco dataset. (a) PauliRGB; (b) Reference Class Map; (c) SVM; (d) 2D-CVNN; (e) Wavelet CNN; (f) CV-CNN-SE; (g) 3D-CVNN; Proposed SDF2Net}
 \label{fig:SF_Results}
\end{figure*}

Based on the findings presented in Tables \ref{tab:FL_Results}--\ref{tab:ober_Results}, it is evident that the proposed SDF2Net model surpasses other methods in performance. Examining the datasets utilized in this study, the Flevoland dataset is composed of 207,832 labeled samples, with 1\% (2,078 samples) reserved for training, spanning across 15 classes. In contrast, the San Francisco dataset boasts a larger pool of labeled samples, with 8,023 allocated for training, explaining the higher classification performance observed in San Francisco. The Oberpfaffenhofen dataset, with 1,311,618 labeled samples and 1\% (13,116 samples) earmarked for training, is characterized by exclusively featuring 3 classes.

The results on Flevoland dataset are reported in Table \ref{tab:FL_Results}, SVM demonstrated the least overall accuracy, primarily because it heavily depends on 1-dimensional information. Additionally, classes with a scant number of training samples, such as Grass, Bare Soil, Wheat 2, Barley, and Buildings, were hardly detected by SVM. This suggests that SVM is unsuitable for datasets with a limited number of training samples, a limitation evident in the accuracies associated with each class. The accuracy of the 2D-CVNN exhibited an enhancement, approximately 10\% greater than that of SVM. Unlike SVM, the 2D-CVNN utilizes 2D filters for spatial information extraction, resulting in a moderate increase in accuracy. Nevertheless, certain classes with a limited number of training samples, like Grass and Bare Soil, recorded lower accuracies, influencing the overall performance of the model.
The Wavelet CNN achieved substantial improvement to the accuracy with 91.73\%, utilizing Wavelet decomposition for feature extraction, thereby enhancing classification accuracy. CV-CNN-SE, employing three parallel branches of two-dimensional kernels, demonstrated an improvement over Wavelet CNN. The 3D-CVNN, using 3D filters for three-dimensional information extraction, did not surpass CV-CNN-SE in accuracy, but did outperform other methods used in this research. Our proposed SDF2Net method outperformed CV-CNN-SE by approximately 1.5\% in overall accuracy and outperformed other methods in seven categories, as detailed in the table \ref{tab:FL_Results}. 
For a visual representation, Fig. \ref{fig:FL_Results} illustrates the classification results of the six methods alongside the reference map. It is clearly shown from Fig. \ref{fig:FL_Results}(c) that SVM had numerous incorrectly assigned pixels, while the classification map generated by SDF2Net closely aligned with the reference map, showcasing superior performance.

\begin{table*}[!t]
\centering
\caption{Experimental Results of different methods on Oberpfaffenhofen Dataset.}
\label{tab:ober_Results}
\begin{tabular}{ccccccccc}
\hline
Class          & Train & Test   & SVM                       & 2D-CVNN          & Wavelet CNN      & CV-CNN-SE           & 3D-CVNN          & SDF2Net                   \\ \hline
Build-Up Areas & 3281  & 324770 & 56.33 & 89.35            & \textbf{93.09}   & 90.49           & 92.58            & 91.01                     \\
Wood Land      & 2467  & 244206 & 57.21 & 92.24   & 90.73            & 96.44            & 94.99            & \textbf{96.80}                     \\
Open Areas     & 7369  & 729525 & 95.98 & 96.20   & 94.10            & 96.14           & 94.43            & \textbf{96.71}                     \\ \hline
OA (\%)        &       &        & 80.46 $\pm$ 1.29          & 94.35 $\pm$ 0.72 & 94.21 $\pm$ 0.49 &  94.86 $\pm$ 0.24 & 94.33 $\pm$ 0.55 & \textbf{95.30 $\pm$ 0.08} \\
AA (\%)        &       &        & 70.84 $\pm$ 2.10          & 93.01 $\pm$ 2.92 & 93.97 $\pm$ 1.47 & 94.49 $\pm$ 0.31 & 94.32 $\pm$ 1.27 & \textbf{94.84 $\pm$ 0.08} \\
k $\times$ 100 &       &        & 64.20 $\pm$ 2.71          & 90.31 $\pm$ 3.64 & 90.26 $\pm$ 1.61 & 91.25 $\pm$ 0.30 & 90.41 $\pm$ 1.43 & \textbf{91.99 $\pm$ 0.13} \\ \hline
\end{tabular}
\end{table*}
\begin{figure*}[t!]
\centering
\includegraphics[width=.95\linewidth]{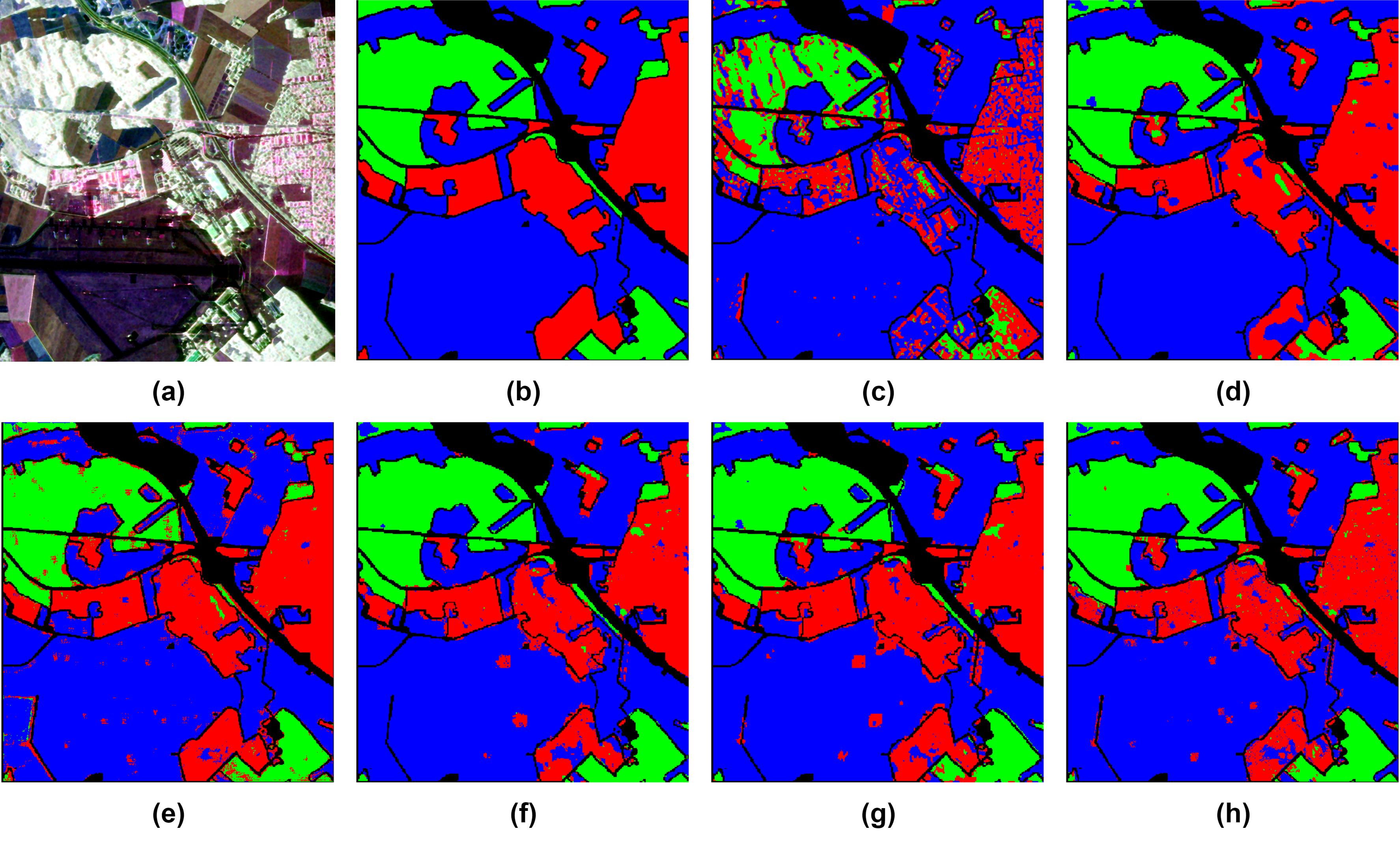}    
 \caption{Classification results of the Oberpfaffenhofen dataset. (a) PauliRGB; (b) Reference Class Map; (c) SVM; (d) 2D-CVNN; (e) Wavelet CNN; (f) CV-CNN-SE; (g) 3D-CVNN; Proposed SDF2Net}
 \label{fig:Ober_Results}
\end{figure*}

Regarding the San Francisco dataset, Table \ref{tab:SF_Results} displays the classification outcomes obtained from various models. In contrast to the Flevoland dataset, this dataset comprises only five target categories. SVM yields relatively poor classification results, suggesting that the original polarimetric features lack effective discrimination. The models 2D-CVNN and CV-CNN-SE show unsatisfactory performance on this dataset, particularly in the Bare Soil category. Conversely, the Wavelet CNN and 3D-CVNN methods prove more adept at handling the intricate polarimetric features, demonstrating superior performance, especially in the Bare Soil category. SDF2Net not only achieves the highest Overall Accuracy (OA) at 97.13\%, Average Accuracy (AA) at 91.31\%, and Kappa coefficient at 95.50\% but also surpasses the performance of other compared methods. In Fig. \ref{fig:SF_Results}, the classification maps generated by the methods employed in this research are presented alongside the reference map. The classification maps produced by our method closely align with the ground-truth map. Especially when looking the the Urban class (Blue), our method is the closer to the reference map when compared with others.

The classification results of Oberpfaffenhofen dataset are shown in Table \ref{tab:ober_Results}. Our proposed method was the highest in terms of OA, AA and Kappa when compared to the other methods. The model had the highest accuracy in all classes. Fig. \ref{fig:Ober_Results} shows the classification results of the methods used in this research. it is very clear that our proposed methods has the closest classification map to the reference data.

According to the experimental findings, the SDF2Net model, as proposed in this study, outperforms other classification methods employed. SVM exhibits the poorest classification performance due to its reliance on 1-dimensional features, leading to a loss of spatial information. In contrast, 2D-CVNN, Wavelet CNN, and CV-CNN-SE consider spatial information, resulting in enhanced overall accuracy compared to SVM. The 3D-CVNN method extracts hierarchical features in both spatial and scattering dimensions through 3-D Complex-Valued convolutions, effectively capturing physical properties from polarimetric adjacent resolution cells. Our proposed SDF2Net demonstrates superior results across the three datasets utilized in this research, leveraging the strengths of each branch to extract features at various levels and thereby enhancing classification performance.

\subsection{Post-Processing with Median Filtering}
To boost the classification accuracy, an additional spatial post-processing stage employs a $3\times3$ median filter to eliminate isolated misclassified pixels within the assigned class. The underlying assumption is that information classes tend to occupy spatial regions of relatively uniform characteristics, typically larger than a few pixels \cite{alkhatib2019improved}. The application of median filtering yields a more refined version of the class map with smoother transitions. To showcase the efficacy of this process, the smoothed classification map is compared with the reference map through the confusion matrix. For illustrative purposes, the Oberpfaffenhofen dataset will be utilized as an exemplar.

\begin{figure*}[t!]
\centering
\includegraphics[width=.95\linewidth]{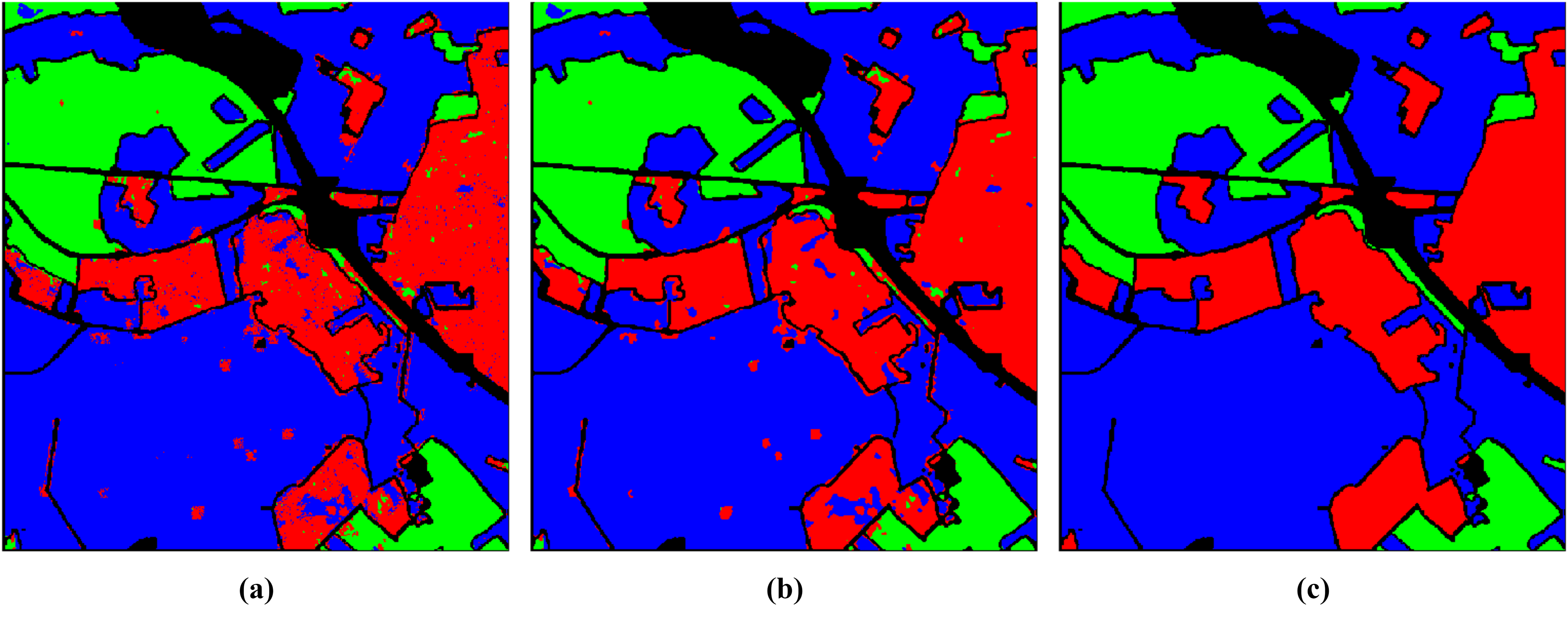}    
 \caption{Urban Image: (a) classification map resulted from the proposed model; (b) classification map after median filtering; (c) reference data classification map}
 \label{fig:Ober_Results_median}
\end{figure*}

\begin{table}[t!]
\centering
\caption{Confusion matrix between reference and generated classification maps for Oberpfaffenhofen dataset.}
\label{tab:ober_conf}
\begin{tabular}{ccccc|}
\cline{3-5}
                                                                                                     & \multicolumn{1}{c|}{}                                                                   & \multicolumn{3}{c|}{\textbf{Reference Data}}                                                                                                                                                                                                  \\ \cline{3-5} 
                                                                                                     & \multicolumn{1}{c|}{}                                                                   & \multicolumn{1}{c|}{\textbf{\begin{tabular}[c]{@{}c@{}}Build-Up \\ Areas\end{tabular}}} & \multicolumn{1}{c|}{\textbf{\begin{tabular}[c]{@{}c@{}}Wood \\ Land\end{tabular}}} & \textbf{\begin{tabular}[c]{@{}c@{}}Open \\ Areas\end{tabular}} \\ \hline
\multicolumn{1}{|c|}{\multirow{3}{*}{\textbf{\begin{tabular}[c]{@{}c@{}}Class \\ Map\end{tabular}}}} & \multicolumn{1}{c|}{\textbf{\begin{tabular}[c]{@{}c@{}}Build-Up \\ Areas\end{tabular}}} & \multicolumn{1}{c|}{297932}                                                             & \multicolumn{1}{c|}{6808}                                                          & 22640                                                          \\ \cline{2-5} 
\multicolumn{1}{|c|}{}                                                                               & \multicolumn{1}{c|}{\textbf{\begin{tabular}[c]{@{}c@{}}Wood \\ Land\end{tabular}}}      & \multicolumn{1}{c|}{6376}                                                               & \multicolumn{1}{c|}{238968}                                                        & 1536                                                           \\ \cline{2-5} 
\multicolumn{1}{|c|}{}                                                                               & \multicolumn{1}{c|}{\textbf{\begin{tabular}[c]{@{}c@{}}Open \\ Areas\end{tabular}}}     & \multicolumn{1}{c|}{23776}                                                              & \multicolumn{1}{c|}{436}                                                           & 712400                                                         \\ \hline
                                                     
\multicolumn{5}{|c|}{\textbf{OA = 95.30\%; AA = 94.83\%; k = 91.99}}                                                                                                                                                                                                                                                                                                                                                                           \\ \hline
\end{tabular}
\end{table}
\begin{table}[t!]
\centering
\caption{Confusion matrix between reference and filtered classification maps for Oberpfaffenhofen dataset.}
\label{tab:ober_conf_median}
\begin{tabular}{ccccc|}
\cline{3-5}
                                                                                                     & \multicolumn{1}{c|}{}                                                                   & \multicolumn{3}{c|}{\textbf{Reference Data}}                                                                                                                                                                                                  \\ \cline{3-5} 
                                                                                                     & \multicolumn{1}{c|}{}                                                                   & \multicolumn{1}{c|}{\textbf{\begin{tabular}[c]{@{}c@{}}Build-Up \\ Areas\end{tabular}}} & \multicolumn{1}{c|}{\textbf{\begin{tabular}[c]{@{}c@{}}Wood \\ Land\end{tabular}}} & \textbf{\begin{tabular}[c]{@{}c@{}}Open \\ Areas\end{tabular}} \\ \hline
\multicolumn{1}{|c|}{\multirow{3}{*}{\textbf{\begin{tabular}[c]{@{}c@{}}Class \\ Map\end{tabular}}}} & \multicolumn{1}{c|}{\textbf{\begin{tabular}[c]{@{}c@{}}Build-Up \\ Areas\end{tabular}}} & \multicolumn{1}{c|}{306268}                                                             & \multicolumn{1}{c|}{4420}                                                          & 15780                                                          \\ \cline{2-5} 
\multicolumn{1}{|c|}{}                                                                               & \multicolumn{1}{c|}{\textbf{\begin{tabular}[c]{@{}c@{}}Wood \\ Land\end{tabular}}}      & \multicolumn{1}{c|}{6132}                                                               & \multicolumn{1}{c|}{238212}                                                        & 1264                                                           \\ \cline{2-5} 
\multicolumn{1}{|c|}{}                                                                               & \multicolumn{1}{c|}{\textbf{\begin{tabular}[c]{@{}c@{}}Open \\ Areas\end{tabular}}}     & \multicolumn{1}{c|}{22576}                                                              & \multicolumn{1}{c|}{296}                                                           & 710360                                                         \\ \hline

\multicolumn{5}{|c|}{\textbf{OA = 96.13\%; AA = 95.49\%; k = 93.42}}                                                                                                                                                                                                                                                                                                                                                                           \\ \hline
\end{tabular}
\end{table}
\begin{table}[t!]
\centering
\caption{Impact of Median Filtering on the classification maps of different datasets}
\label{tab:med_summary}
\begin{tabular}{l|cc|cc|cc|}
\cline{2-7}
                                              & \multicolumn{2}{c|}{\textbf{Flevoland}} & \multicolumn{2}{c|}{\textbf{San Francisco}} & \multicolumn{2}{c|}{\textbf{Oberpfaffenhofen}} \\ \cline{2-7} 
                                              & \multicolumn{1}{c|}{Before}   & After   & \multicolumn{1}{c|}{Before}     & After     & \multicolumn{1}{c|}{Before}       & After      \\ \hline
\multicolumn{1}{|l|}{\textbf{OA (\%)}}        & \multicolumn{1}{c|}{96.01}    & 96.82   & \multicolumn{1}{c|}{97.13}      & 97.60     & \multicolumn{1}{c|}{95.30}        & 96.13      \\ \hline
\multicolumn{1}{|l|}{\textbf{AA (\%)}}        & \multicolumn{1}{c|}{95.17}    & 95.59   & \multicolumn{1}{c|}{91.31}      & 91.84     & \multicolumn{1}{c|}{94.84}        & 95.49      \\ \hline
\multicolumn{1}{|l|}{\textbf{k $\times$ 100}} & \multicolumn{1}{c|}{95.64}    & 96.52   & \multicolumn{1}{c|}{95.50}      & 96.22     & \multicolumn{1}{c|}{91.99}        & 93.42      \\ \hline
\end{tabular}
\end{table}
Fig. \ref{fig:Ober_Results_median} displays classification maps for Oberpfaffenhofen. In Fig. \ref{fig:Ober_Results_median}(a), the map is generated using the proposed model. Fig. \ref{fig:Ober_Results_median}(b) depicts the same map after smoothing with a median filter. Fig. \ref{fig:Ober_Results_median}(c) represents the reference data classification map. Table \ref{tab:ober_conf} exhibits the confusion matrix comparing the generated classification map with the reference, while Table \ref{tab:ober_conf_median} shows the confusion matrix for the filtered generated classification map in comparison to the reference.

The diagonal of the confusion matrix demonstrates the agreement between the predicted class map and the reference class map, while discrepancies are indicated by the non-diagonal elements. Tables \ref{tab:ober_conf} and \ref{tab:ober_conf_median} display a decline in the values of these non-diagonal components, signaling a decrease in the number of misclassified pixels. As a result, there is an enhancement in Overall Accuracy. This improvement is noticeable, particularly in the Build-Up Areas class (depicted in red). Fig. \ref{fig:Ober_Results_median}(a) highlights numerous pixels initially classified as Open Areas. However, a substantial correction is observed after applying the median filter. This correction is corroborated by Tables \ref{tab:ober_conf} and \ref{tab:ober_conf_median}, revealing an initial misclassification of 22,640 Build-Up Areas pixels as Open Areas, which reduces to 15,780 after the application of the median filter. In fact, the reduction appears on all non-diagonal values. Table \ref{tab:med_summary} provides a summary of the enhancement achieved by applying the median filter to the resulting class map of each dataset.

\subsection{Performance of Different Models at Different Percentages of Training Data}
\begin{figure*}[tb]
\centering
\includegraphics[width=.99\linewidth]{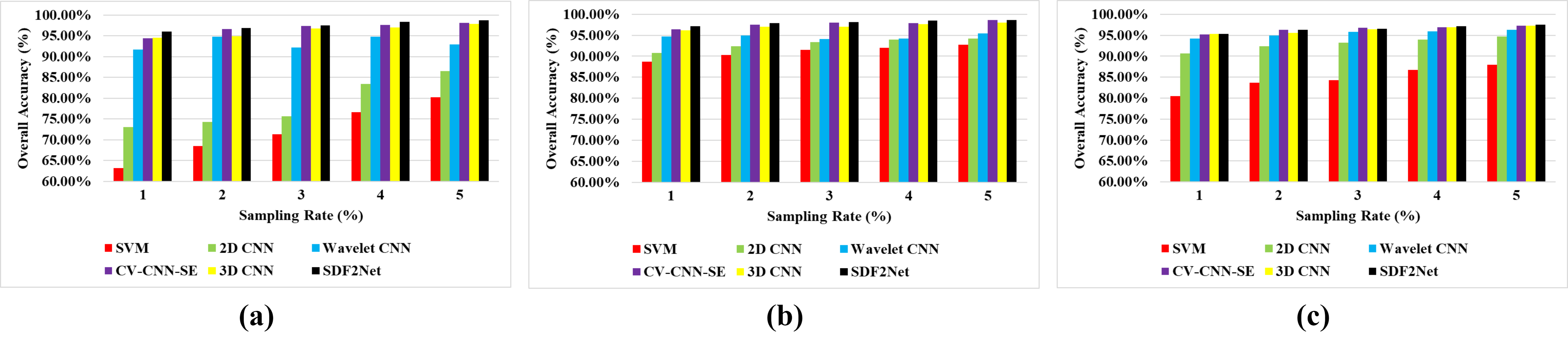}    
 \caption{Classification accuracy at different percentages of training data (a) Flevoland; (b) San Francisco; (c) Oberpfaffenhofen.}
 \label{fig:sampling_Results}
\end{figure*}

The model's performance can be effectively assessed by examining the classification accuracy across different percentages of training data. We randomly chose 1\%, 2\%, 3\%, 4\%, and 5\% of labeled samples for training, using the remaining samples for testing. The classification results for each dataset are depicted in Fig. \ref{fig:sampling_Results}. It is evident that, across all methods employed in this study, the classification accuracy shows improvement with an increase in the number of training samples. Notably, the proposed model consistently outperforms others across all proportions of training samples on the three datasets.

\section{Conclusion}
\label{sec:conclusion}
This paper introduces a novel model called SDF2Net designed for PolSAR image classification. The model addresses existing challenges in PolSAR classification by incorporating a three-branch feature fusion structure and optimizing the creation of a complex-valued CNN-based model. The data is processed through three branches of CV-3D-CNN, and the generated features from each branch are combined. The fused features undergo enhancement through an attention block to improve model performance, followed by the application of fully connected and dropout layers to yield the final classification result. Experimental results demonstrate the effectiveness of the proposed model in terms of OA, AA, and Kappa. Notably, even with limited training data, the model produces classification results almost identical to the reference data.

As part of future work, we plan to explore a lighter architecture with fewer training parameters to reduce computational complexity without compromising model performance. Additionally, we aim to investigate a more compact neural network model that enhances generalization ability and achieves satisfactory classification accuracy across diverse datasets.

\bibliographystyle{IEEEtran}
\bibliography{bibtex/bib/Refs}

\begin{IEEEbiography}[{\includegraphics[width=1in,height=1.25in,clip,keepaspectratio]{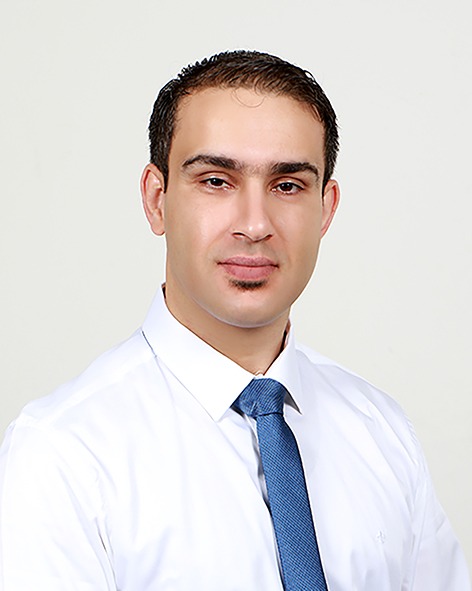}}]{Mohammed Q. Alkhatib} (S'09, M'18, SM'24) earned his B.S. degree in Telecommunications Engineering from Yarmouk University, Irbid, Jordan, in 2008. Subsequently, he completed his M.S. and Ph.D. degrees in Electrical and Computer Engineering at the University of Texas at El Paso, El Paso, TX, USA, in 2011 and 2018, respectively. 
 
 Between 2014 and 2020, Dr. Alkhatib served as a lecturer at Abu Dhabi Polytechnic, Abu Dhabi, UAE. Currently, he holds the position of Assistant Professor at the College of Engineering and IT in the University of Dubai, Dubai, UAE. His research interests are centered around remote sensing and machine learning.
 \end{IEEEbiography}

\vskip -2\baselineskip plus -1fil

\begin{IEEEbiography}[{\includegraphics[width=1in,height=1.25in,clip,keepaspectratio]{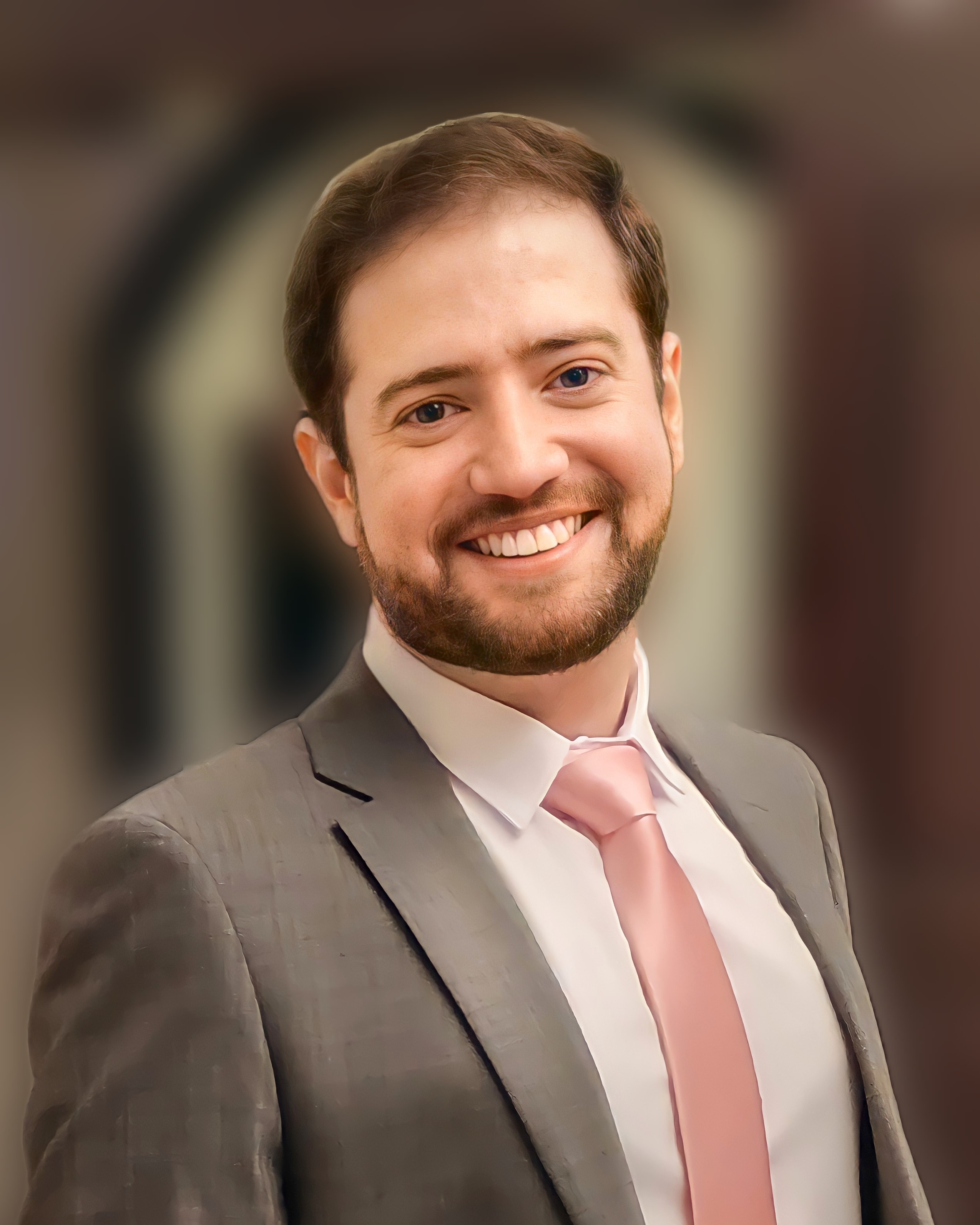}}]
{M. Sami Zitouni}
received his PhD and MSc degrees in Electrical and Computer Engineering in 2019 and 2015, respectively, from Khalifa University (KU), Abu Dhabi, UAE. He conducted his studies with the KU Center for Autonomous Robotic Systems (KUCARS) and the Visual Signal Analysis and Processing Center (VSAP) in topics including video processing and analysis, crowd modeling, detection and tracking, social and cognitive behavior analysis, and visual scene understanding. Before joining the University of Dubai as Assistant Professor, he was a Post-Doctoral Fellow in Biomedical Engineering at KU as part of KU – Korean Advanced Institute of Science and Technology (KU-KAIST) Joint Research Center where he worked on physiological signals analysis, affective state recognition, mental health monitoring, and emotionally aware human-machine interaction. Currently, he is in charge of Mohammed Bin Rashid Space Center (MBRSC) Lab. His research interests include artificial intelligence, machine learning applications, remote sensing, earth observation, affective computing, health informatics, computer vision, and embedded systems.
\end{IEEEbiography}

\vskip -2\baselineskip plus -1fil

\begin{IEEEbiography}[{\includegraphics[width=1in,height=1.25in,clip,keepaspectratio]{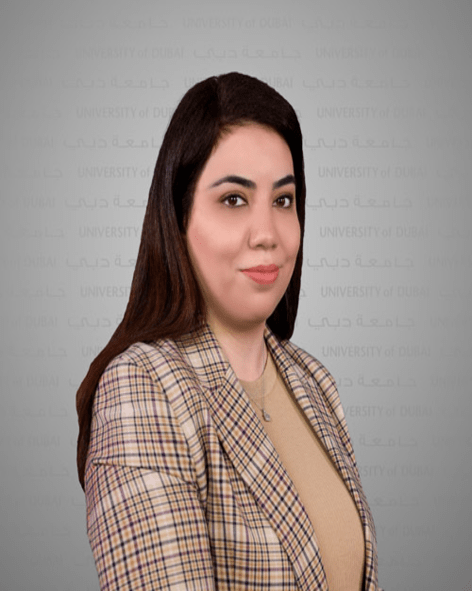}}]
{Mina Al-Saad}
received her BSc in Laser and Optoelectronics Engineering from Al-Nahrain University, Baghdad, Iraq, in 2009, followed by her MSc in the same field from the same university in 2012. Since then, she has held the position of Research Associate at the Mohammed Bin Rashid Space Centre (MBRSC) Lab, University of Dubai, actively contributing to various projects. Prior to this role, she served as a GIS engineer at the Ministry of Construction and Housing in Baghdad, Iraq, from 2013 to 2017. Her research interests include Geographic Information System (GIS) and remote sensing, focusing on integrating Artificial Intelligence (AI) technologies into remote sensing applications for tasks such as classification, autonomous object detection, semantic segmentation, as well as satellite calibration and validation activities.

\end{IEEEbiography}

\begin{IEEEbiography}[{\includegraphics[width=1in,height=1.25in,clip,keepaspectratio]{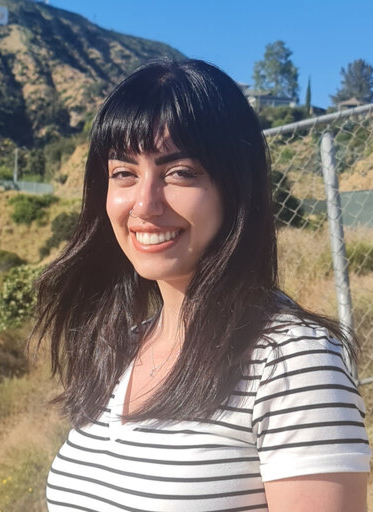}}]
{Nour Aburaed} (M'18) received her Ph.D. in Electronic and Electrical Engineering from the University of Strathclyde (Glasgow, UK) in 2023, which is specialized in Single Image Super Resolution for spatial enhancement of hyperspectral remote sensing imagery. Nour received her M.Sc. degree in Electrical and Computer Engineering from Khalifa University of Science and Technology (Abu Dhabi, UAE) in 2016, particularly specialized in High-ISO image de-noising and Quantum Image Processing.\\
\indent Between 2016 and 2018, Nour served as a Teaching Assistant at Khalifa University, engaging in a wide range of theoretical and laboratory-based courses, such as Calculus and Physics. Her career took a significant turn in June 2018 when she joined the Mohammed Bin Rashid Space Centre (MBRSC) Laboratory, situated at the University of Dubai. Here, she initially served as a Research Assistant (Jun. 2018 - Sep. 2023) before being elevated to the position of Research Associate (Sep. 2023 - current). Nour's expertise in applying image processing and artificial intelligence within the context of remote sensing has yielded various research findings that are utilized by MBRSC and published in reputable conferences and journals.
\indent Nour is an active member of IEEE; she serves as the secretary of IEEE ComSoc (UAE Chapter). She also actively volunteers by serving as a chair and a reviewer in various conferences and journals (ICASSP, IGARSS, ICIP, JSTARS, TGRS, and MDPI Remote Sensing). Nour's main research interests include Hyperspectral Imagery, Super Resolution, Object Detection, Semantic Segmentation, Convolutional Neural Networks, Domain Adaptation, and Satellite Calibration and Validation.
\indent Nour was the recipient of the President's Scholarship and Master Research Teaching Scholarship (MRTS) from Khalifa University of Science and Technology for International Students.
\end{IEEEbiography}

\begin{IEEEbiography}[{\includegraphics[width=1in,height=1.25in,clip,keepaspectratio]{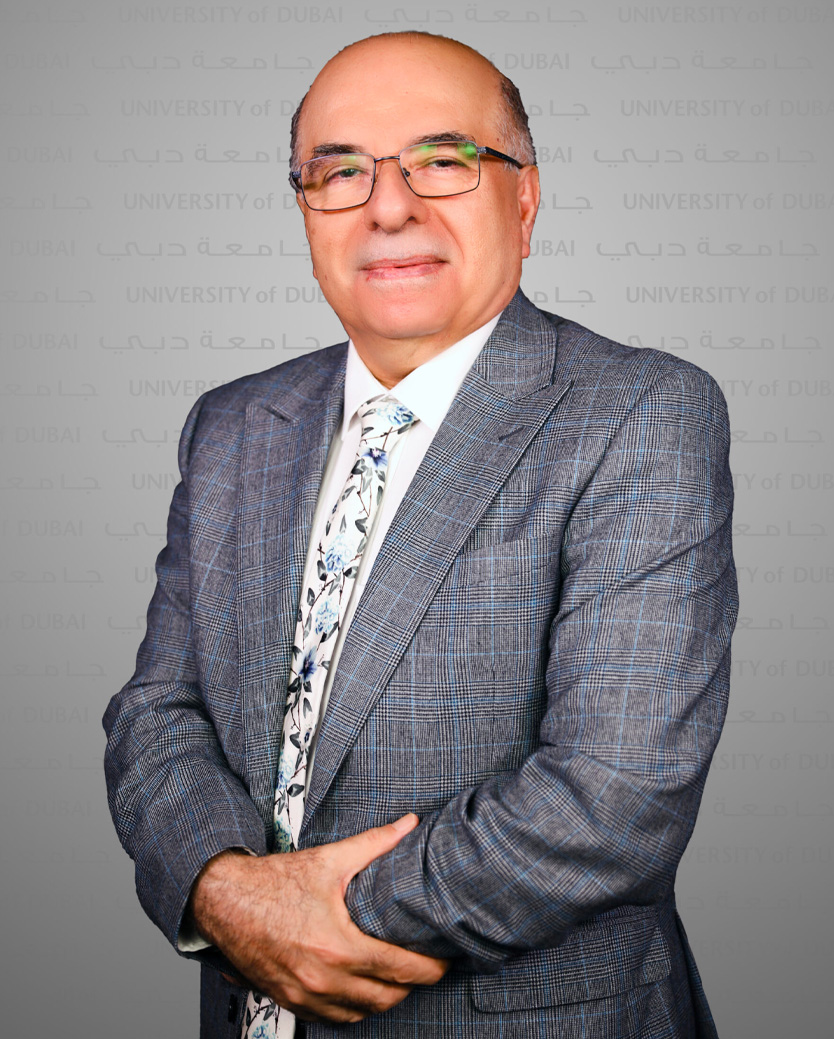}}]
{Hussain Al Ahmad}
 (S'78, M'83, SM'90, LSM'20) has a Ph.D. from the University of Leeds, UK in 1984 and currently he is the Provost and Chief Academic Officer at the University of Dubai, UAE. He has 37 years of higher education experience working at academic institutions in different countries including University of Portsmouth, UK, Leeds Beckett University, UK, Faculty of Technological Studies, Kuwait, University of Bradford, UK, Etisalat University College, Khalifa University and University of Dubai, UAE. He is the founding Dean of Engineering and IT at the University of Dubai, UAE. He is the founder and Chair of the Electronic Engineering department at both Khalifa University and Etisalat University College. His research interests are in the areas of signal and image processing, artificial intelligence, remote sensing and propagation. He has supervised successfully 32 PhD and Master students in the UK and UAE. He has delivered short courses and seminars in Europe, Middle East and Korea. He has published 120 papers in international conferences and journals. He has UK and US patents. He served as chairman and member of the technical program committees of many international conferences. He is a Life Senior Member of the IEEE, a Fellow of the Institution of Engineering and Technology (FIET), Chartered Engineer (C.Eng), Fellow of the British Royal Photographic Society (FRPS) and Accredited Senior Imaging Scientist (ASIS). 
\end{IEEEbiography}

\end{document}